\documentclass{article}

\usepackage{pifont}
\usepackage{booktabs}
\usepackage{multirow}
\usepackage{xcolor}
\newcommand{\cmark}{\ding{51}}
\newcommand{\xmark}{\ding{55}}
\usepackage[utf8]{inputenc}
\usepackage[T1]{fontenc}
\usepackage{hyperref}
\usepackage{url}
\usepackage{booktabs}
\usepackage{amsfonts}
\usepackage{amsmath}
\usepackage{nicefrac}
\usepackage{microtype}
\usepackage{xcolor}
\usepackage{tabularx}
\usepackage{graphicx}
\usepackage{todonotes}
\usepackage[preprint]{corl_2026}

\title{EmbodimentSemantic: A Spatial Scene-Graph Dataset and Benchmark for Vision-Language Models on Embodied Manipulation Trajectories}

\author{
    Hassan Jaber$^{1}$, Refinath S N$^2$, Luca Cagliero$^1$,\\
    {\bfseries Christopher E. Mower$^2$, Haitham Bou-Ammar$^{2, 3}$}\\
    $^1$Politecnico di Torino, Italy,\\
    $^2$Huawei, Noah's Ark Lab, United Kingdom,\\
    $^3$University College London, United Kingdom
}

\begin{document}

\maketitle
\begin{abstract}
Spatial grounding remains a key limitation of vision--language--action (VLA) systems for robotic manipulation. While current models can recognize objects and follow language instructions, they often lack an explicit representation of how objects are arranged in space, including support, containment, ordering, occlusion, and depth-sensitive relations. We introduce \textsc{EmbodimentSemantic}, a spatial scene-graph dataset and benchmark for evaluating relational grounding in embodied manipulation. \textsc{EmbodimentSemantic} represents scenes as directed object--relation--object triplets, where each triplet specifies a spatial relation between an ordered pair of objects using a fixed set of relations. This representation enables direct evaluation of object binding, relation prediction, and spatial consistency. The dataset includes real-world manipulation observations collected with the low-cost SO101 robot arm, together with generated scene graphs for studying spatial grounding in practical robotic settings. To provide controlled validation, we also introduce a simulator-grounded LIBERO benchmark with over 60K manipulation frames and more than 120K camera-specific scene graphs across paired third-person and wrist views, where ground-truth relations are derived automatically from MuJoCo geometry, world coordinates, camera projections, and visibility constraints. We further test whether scene graphs improve downstream control by injecting them into existing VLA policy prompts. Experiments across open-source and commercial VLMs show that current models often predict plausible relations but struggle with exact depth-aware and viewpoint-dependent spatial structure. \textsc{EmbodimentSemantic} provides a unified framework for diagnosing spatial grounding in VLM perception and testing its utility for VLA manipulation.

\end{abstract}
\begin{figure}[t]
    \centering
    \includegraphics[width=1\linewidth]{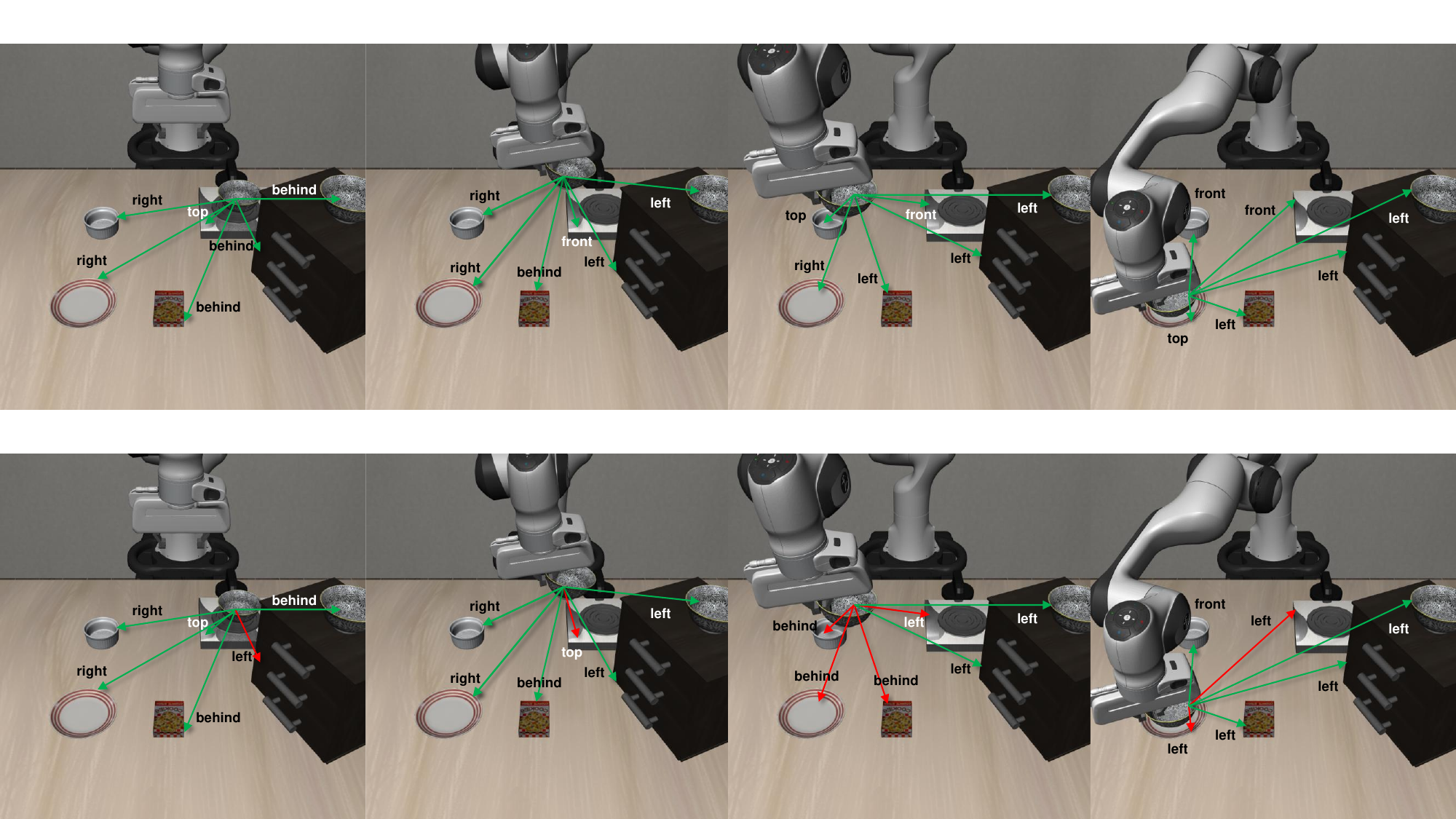}
    \caption{
LIBERO scene-graph comparison. 
The top row shows simulator-grounded ground-truth relations, and the bottom row shows \texttt{gemini-3.1-pro} predictions for the same frames. 
Green edges indicate correct triplets and red edges indicate errors, highlighting both successful relation recovery and remaining depth-sensitive failures.
}
    \label{fig:vlm-qualitative}
\end{figure}

\section{Introduction}

Since the development of the transformer architecture~\cite{vaswani2017attention}, robot learning has seen several major advances~\cite{pmlr-v205-ichter23a, pi0}. 
Recent vision--language--action (VLA) models extend this trend by adapting internet-scale vision--language models (VLMs) to robot control, typically through fine-tuning on robot demonstrations~\cite{zitkovich2023rt2,kim2025openvla}. 
Despite their scale, however, these models remain weakly grounded in scene geometry. 
Several works show that VLA policies can learn shortcut associations between actions and task-irrelevant visual or semantic context, such as background, texture, viewpoint, familiar objects, or linguistic priors, rather than reliably representing spatial relations such as distance, relative pose, object size, and viewpoint~\cite{xing2025shortcut,zhang2025inspire,fang2025intention}.

Recent work has begun to address this gap by injecting spatial supervision and structured spatial representations into VLMs and VLAs, including large-scale metric spatial VQA data, robot-relevant annotations from 3D scans, and explicit egocentric spatial encodings or action grids~\cite{chen2024spatialvlm,song2025robospatial,QuD-RSS-25}. 
However, these methods still largely rely on learned spatial priors and do not remove the fundamental ambiguity of inferring metric geometry from RGB appearance alone. 
For example, if one object has a larger image footprint than another, it may be physically larger, closer to the camera, or both. 
This size--distance ambiguity means that appearance alone does not uniquely determine metric scale, depth, or pose without additional geometric cues such as depth, stereo, motion, camera/object pose, or object-size priors~\cite{battaglia2011haptic}.

Earlier robotic perception work addressed geometry more explicitly through object-pose estimation from RGB or RGB-D observations~\cite{xiang2018posecnn,tremblay2018deep,hodan2018bop}, and such estimates have long supported manipulation through visual servoing~\cite{hutchinson1996tutorial,chaumette2006visual}. 
More recent manipulation policies similarly explore explicit spatial structure by augmenting RGB-language inputs with representations such as 3D keypoints, predicted end-effector poses, point clouds, or object-centric trajectories~\cite{shridhar2023peract,goyal2023rvt,li2024manipllm,li2025language,hsu2025spot,li20253dsvla}. 
While these representations make geometry more directly available to the policy, they often shift the burden onto the perception system, requiring reliable depth sensing, calibration, segmentation, pose tracking, or related infrastructure, which can limit generalization.

Another challenge is evaluation: success on spatial reasoning benchmarks does not necessarily imply recovery of a structured scene representation. 
Existing multimodal benchmarks primarily evaluate spatial understanding through captioning, visual question answering, relation classification, or image--text matching~\citep{krishna2017visualgenome, yang2019spatialsense, liu2023vsr, thrush2022winoground, wang2024spatialeval, chen2024spatialvlm}. 
Although useful for diagnosing spatial and compositional failures, these evaluations typically produce free-form answers, binary judgments, or matching scores rather than explicit relational states. 
As a result, they do not directly measure whether a model can recover the structured representations needed for robotic planning, monitoring, or action conditioning.

We introduce \textsc{EmbodimentSemantic}, a spatial scene-graph dataset and benchmark for evaluating VLMs on embodied manipulation trajectories. 
Our primary contribution is a real-world dataset collected with the low-cost SO101 robot arm, pairing manipulation observations with generated scene graphs that encode object-centric spatial relations as directed triplets $(\texttt{objectA}, \texttt{relation}, \texttt{objectB})$. 
To validate scene-graph generation under controlled conditions, we also construct a simulator-grounded LIBERO~\cite{liu2023libero} benchmark in which ground-truth scene graphs are generated from simulator state and compared against VLM-predicted graphs from robot camera observations. 
Using both real-world and simulated data, we benchmark open-source and commercial VLMs on their ability to recover manipulation-relevant spatial relations. 
Finally, we study whether predicted scene graphs can improve downstream robot control by injecting them into existing VLA policy prompts, in the spirit of composing pretrained models through intermediate language representations~\cite{zeng2023socratic}. 
This scene-graph augmentation improves performance on several task-policy pairs without additional VLA fine-tuning, while also revealing cases where added relational context can hurt policy behavior, suggesting that explicit relational context can help bridge VLM spatial understanding and VLA action generation.

This paper introduces \textsc{EmbodimentSemantic}, a spatial scene-graph dataset for embodied manipulation trajectories, with a primary real-world component collected using the low-cost SO101 robot arm. The dataset pairs robot observations with object-centric scene graphs represented as directed relational triplets, defined under a fixed bidirectional relation ontology. Alongside the real-world data, we develop a simulator-grounded LIBERO benchmark for studying scene-graph generation in controlled settings. Using MuJoCo object geometry, world coordinates, and projected 2D bounding boxes from robot camera views, we produce frame-level ground-truth scene graphs and use them to compare against graphs predicted by vision-language models. We also evaluate open-source and commercial VLMs on structured triplet prediction through constrained prompting and triplet-level metrics. Finally, we study how these generated graphs can support robot control by injecting them into existing fine-tuned VLA policies, enabling analysis of policy robustness under scene perturbations, viewpoint shifts, object randomization, and prompt modifications within the LeRobot evaluation pipeline.

\section{Related work}

\paragraph{Scene graphs and spatial reasoning benchmarks.}
Scene graphs provide a structured way to represent objects and their relations, and have been widely used to study relational visual understanding. Visual Genome~\citep{krishna2017visualgenome} introduced large-scale object, attribute, and relationship annotations for natural images, while SpatialSense~\citep{yang2019spatialsense}, Visual Spatial Reasoning (VSR)~\citep{liu2023vsr}, and Winoground~\citep{thrush2022winoground} evaluate spatial or compositional grounding through relation classification, image--text pairs, and caption matching. More recent VLM benchmarks further target spatial reasoning: SpatialEval~\citep{wang2024spatialeval} evaluates relationship reasoning, counting, navigation, and positional understanding; SpatialVLM~\citep{chen2024spatialvlm} trains on large-scale metric spatial question-answering data; SpatialRGPT~\citep{cheng2024spatialrgpt} studies grounded 2D and 3D spatial reasoning; and EmbSpatial-Bench~\citep{du2024embspatial} and MV-RoboBench~\citep{feng2026seeing} move toward embodied or robotic scenes. These benchmarks provide useful diagnostics for spatial reasoning, but they typically evaluate answers, classifications, or matching decisions rather than the full relational structure of a manipulation scene. \textsc{EmbodimentSemantic} instead evaluates whether models can directly produce ordered scene-graph triplets $(\texttt{objectA}, \texttt{relation}, \texttt{objectB})$ from embodied manipulation observations, using a fixed set of spatial relations. This makes it possible to isolate object-binding errors, relation reversals, directional inconsistencies, and hallucinated relations that may be hidden in language-facing benchmarks but can directly change the action required for manipulation.

\paragraph{Embodied learning and VLA systems.}
Robotic learning benchmarks and datasets such as LIBERO~\citep{liu2023libero}, RLBench~\citep{james2020rlbench}, and Open X-Embodiment~\citep{openx2024} support evaluation of manipulation generalization, transfer, and cross-embodiment learning. RT-1~\citep{brohan2022rt1} demonstrated scalable transformer policies trained on large robot datasets, while RT-2~\citep{zitkovich2023rt2}, PaLM-E~\citep{driess2023palme}, and OpenVLA~\citep{kim2025openvla} show how vision-language pretraining and multimodal foundation models can support embodied reasoning, planning, and robot control. These systems are typically evaluated through downstream task success, which does not directly reveal whether the model has recovered the spatial relations present in its observations. \textsc{EmbodimentSemantic} complements these benchmarks by evaluating relational grounding directly and by testing whether explicit scene-graph context improves VLA policy behavior.

\paragraph{Spatial structure in robot policies.}
A growing body of work incorporates explicit spatial structure into robot policies. PerAct~\citep{shridhar2023peract} and RVT~\citep{goyal2023rvt} use 3D voxelized or multi-view representations, while ManipLLM~\citep{li2024manipllm}, LaNO3DP~\citep{li2025language}, SPOT~\citep{hsu2025spot}, and 3DS-VLA~\citep{li20253dsvla} use object-centric, 3D, keypoint, pose, or trajectory-level structure. Recent VLA methods introduce spatial reasoning more directly: SpatialVLA~\citep{QuD-RSS-25} uses spatial position encodings and action representations, InSpire~\citep{zhang2025inspire} adds spatial reasoning questions over task-relevant objects, CoT-VLA~\citep{zhao2025cotvla} reasons through future-frame prediction, and GraphCoT-VLA~\citep{huang2026graphcotvla} incorporates an updatable 3D pose-object graph. 
While effective, these approaches often require policy modifications, additional training, or additional perception infrastructure, such as RGB-D sensing, calibration, or pose tracking.
\textsc{EmbodimentSemantic} is complementary: it generates relational scene graphs from standard visual observations and injects them into existing fine-tuned VLA prompts at evaluation time, requiring no additional sensors, retraining, or architecture-specific changes.

\paragraph{Robustness and annotation extensions for robot data.}
Robustness-oriented LIBERO extensions show that standard task success can obscure brittle policy behavior. LIBERO-PRO~\citep{zhou2026liberoprorobustfairevaluation} evaluates perturbations to objects, initial states, instructions, and environments, while LIBERO-Plus~\citep{fei2025liberoplus} studies factors such as layout, viewpoint, robot state, language, lighting, texture, and sensor noise. Other work enriches robot demonstrations with additional supervision: SlotVLA~\citep{hanyu2025slotvla} introduces LIBERO+ with box, mask, and tracking annotations, and PixelVLA~\citep{liang2026pixelvla} studies pixel-level understanding for VLA models. \textsc{EmbodimentSemantic} is complementary: it focuses on object-centric spatial scene graphs for embodied manipulation, combining a real-world SO101 dataset with a simulator-grounded LIBERO benchmark for controlled VLM triplet evaluation. It further tests whether generated scene graphs improve downstream control by injecting them into existing VLA prompts within the LeRobot evaluation pipeline~\citep{cadene2026lerobot}.

\begin{table}[t]
\centering
\scriptsize
\setlength{\tabcolsep}{3pt}
\renewcommand{\arraystretch}{0.95}
\caption{Comparison with representative spatial reasoning, scene-graph, and robotic manipulation benchmarks. Columns indicate whether a benchmark is LIBERO-based, embodied, scene-graph based, uses directed spatial relations, supports multiple views, provides simulator-grounded labels, evaluates exact object--relation--object triplets, or integrates with VLA policy evaluation.}
\label{tab:prior-work-comparison}
\resizebox{\linewidth}{!}{
\begin{tabular}{lcccccccc}
\toprule
Benchmark 
& LIBERO 
& Emb. 
& SG 
& Dir. rel. 
& Multi-view 
& Sim GT 
& Exact trip. 
& VLA int. \\
\midrule

Visual Genome~\citep{krishna2017visualgenome}
& \xmark & \xmark & \cmark & \cmark & \xmark & \xmark & \xmark & \xmark \\

SpatialSense~\citep{yang2019spatialsense}
& \xmark & \xmark & \xmark & \cmark & \xmark & \xmark & \xmark & \xmark \\

VSR~\citep{liu2023vsr}
& \xmark & \xmark & \xmark & \xmark & \xmark & \xmark & \xmark & \xmark \\

SpatialEval~\citep{wang2024spatialeval}
& \xmark & \xmark & \xmark & \xmark & \xmark & \xmark & \xmark & \xmark \\

EmbSpatial-Bench~\citep{du2024embspatial}
& \xmark & \cmark & \xmark & \cmark & \xmark & \xmark & \xmark & \xmark \\

MV-RoboBench~\citep{feng2026seeing}
& \xmark & \cmark & \xmark & \xmark & \cmark & \xmark & \xmark & \xmark \\

LIBERO / LIBERO-Plus~\citep{liu2023libero,fei2025liberoplus}
& \cmark & \cmark & \xmark & \xmark & \cmark & \cmark & \xmark & \cmark \\

SlotVLA / LIBERO+~\citep{hanyu2025slotvla}
& \cmark & \cmark & \xmark & \xmark & \cmark & \cmark & \xmark & \cmark \\

PixelVLA~\citep{liang2026pixelvla}
& \cmark & \cmark & \xmark & \xmark & \xmark & \xmark & \xmark & \cmark \\

\textbf{\textsc{EmbodimentSemantic}}
& \cmark & \cmark & \cmark & \cmark & \cmark & \cmark & \cmark & \cmark \\

\bottomrule
\end{tabular}
}
\end{table}

\section{\textsc{EmbodimentSemantic} benchmark and dataset}

\begin{figure}[t]
    \centering
    \includegraphics[width=0.48\linewidth]{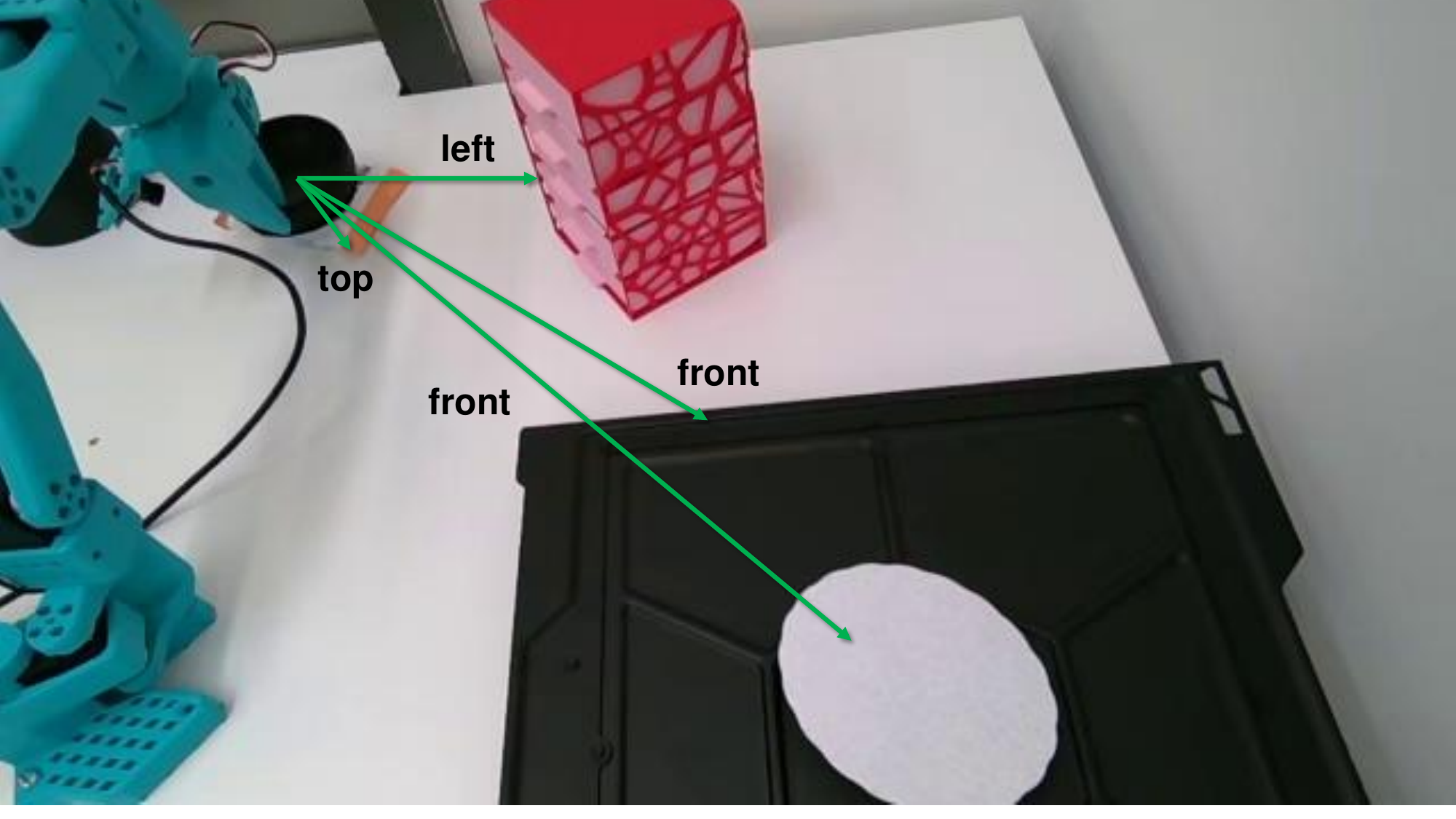}
    \hfill
    \includegraphics[width=0.48\linewidth]{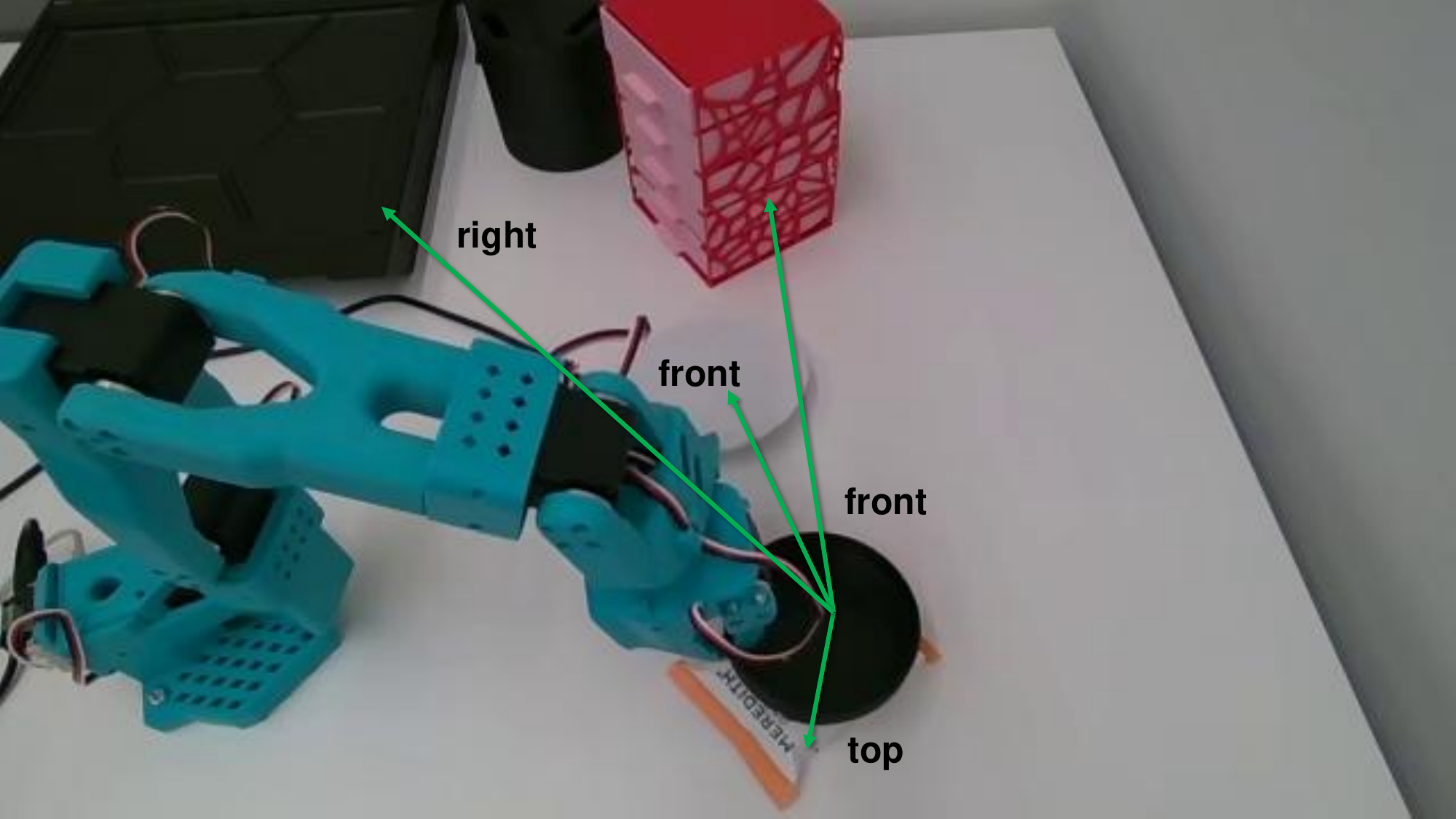}
    \caption{
    Real-world SO101 scene-graph prediction examples.
    Generated scene graphs are overlaid on external-camera observations from two physical robot scenes.
    These examples show that the same object-centric scene-graph representation used in LIBERO can be applied to real SO101 observations.
    }
    \label{fig:so101-scene-graph-examples}
\end{figure}
\textsc{EmbodimentSemantic} is built from LIBERO Spatial manipulation trajectories. Each task consists of demonstrations in which the robot picks up \texttt{akita\_black\_bowl\_1} from a different initial spatial configuration and places it on \texttt{plate\_1}. The benchmark converts these demonstrations into frame-level spatial scene graphs over a fixed object and relation ontology. Each demonstration contains synchronized RGB observations from a fixed third-person camera (\texttt{agentview}) and a wrist-mounted camera (\texttt{eye\_in\_hand}). For each selected frame and camera view, EmbodimentSemantic provides spatial relations as ordered triplets of the form $(\texttt{objectA}, \texttt{relation}, \texttt{objectB})$.

\begin{table}[t]
    \caption{LIBERO Spatial tasks used in EmbodimentSemantic. All tasks share the same goal, placing the black bowl on the plate, but differ in the bowl's initial spatial context.}
    \label{tab:EmbodimentSemantic-tasks}
    \centering
    \begin{tabularx}{\linewidth}{lX}
        \toprule
        Task ID & Initial spatial context of \texttt{akita\_black\_bowl\_1} \\
        \midrule
        T0 & Between the plate and the ramekin \\
        T1 & From table center \\
        T2 & In the top drawer of the wooden cabinet \\
        T3 & Next to the cookie box \\
        T4 & Next to the plate \\
        T5 & Next to the ramekin \\
        T6 & On the cookie box \\
        T7 & On the ramekin \\
        T8 & On the stove \\
        T9 & On the wooden cabinet \\
        \bottomrule
    \end{tabularx}
\end{table}

\paragraph{Object and relation ontology.}
EmbodimentSemantic fixes the object vocabulary and relation set so that model predictions can be evaluated by exact triplet matching. The object vocabulary contains seven canonical LIBERO objects: the manipulated bowl, a distractor bowl, the cookie box, ramekin, plate, wooden cabinet, and stove. The relation vocabulary contains eight directed predicates arranged in four inverse pairs: left/right, front/behind, on-top/below, and inside/contains. This fixed ontology removes ambiguity in object naming and relation formatting, allowing direct comparison across models.

\begin{table}[t]
    \caption{Object and relation ontology used by EmbodimentSemantic. Relations are represented as directed predicates arranged in inverse pairs.}
    \label{tab:EmbodimentSemantic-ontology}
    \centering
    \begin{tabularx}{\linewidth}{llX}
        \toprule
        Type & Name & Description \\
        \midrule
        Object & \texttt{akita\_black\_bowl\_1} & Manipulated bowl instance \\
        Object & \texttt{akita\_black\_bowl\_2} & Secondary bowl instance / distractor \\
        Object & \texttt{cookies\_1} & Cookie box object \\
        Object & \texttt{glazed\_rim\_porcelain\_ramekin\_1} & Ramekin reference object \\
        Object & \texttt{plate\_1} & Target placement object \\
        Object & \texttt{wooden\_cabinet\_1} & Cabinet / drawer reference object \\
        Object & \texttt{flat\_stove\_1} & Stove reference surface \\
        \midrule
        Relation & \texttt{is\_left\_of} / \texttt{is\_right\_of} & Lateral world-frame ordering \\
        Relation & \texttt{is\_in\_front\_of} / \texttt{is\_behind} & Depth world-frame ordering \\
        Relation & \texttt{is\_on\_top\_of} / \texttt{is\_below\_of} & Vertical support or stacking relation \\
        Relation & \texttt{is\_inside} / \texttt{contains} & Containment relation \\
        \bottomrule
    \end{tabularx}
\end{table}

For aggregate analysis, the evaluator can also collapse inverse directions into four relation families: \texttt{is\_left\_of}, \texttt{is\_in\_front\_of}, \texttt{is\_on\_top\_of}, and \texttt{is\_inside}. The primary benchmark, however, evaluates the full directed relation set.

\paragraph{Dataset organization.}
The benchmark contains ten task files, one per LIBERO Spatial task. Each file contains 50 demonstrations, giving 500 demonstrations in total. Demonstration lengths range from 75 to 197 timesteps, with a mean length of 124.5 timesteps. Across all demonstrations, EmbodimentSemantic contains 62,250 paired timesteps and 124,500 RGB frames across the two camera views. RGB observations are stored at $128 \times 128$ resolution.

Frame-level scene graphs are stored under camera-specific annotation keys. The \texttt{agentview} graph describes the third-person view of the scene, while the \texttt{eye\_in\_hand} graph is visibility-filtered for the wrist camera. During annotation generation, EmbodimentSemantic computes MuJoCo world coordinates and projected 2D bounding boxes for the relevant objects; the VLM benchmark evaluates the resulting spatial triplets.

\begin{table}[t]
    \caption{EmbodimentSemantic dataset summary. A paired timestep contains one \texttt{agentview} frame and one \texttt{eye\_in\_hand} frame.}
    \label{tab:EmbodimentSemantic-dataset-summary}
    \centering
    \begin{tabular}{lr}
        \toprule
        Dataset attribute & Value \\
        \midrule
        Tasks & 10 \\
        Demonstrations & 500 \\
        Demonstrations per task & 50 \\
        Paired timesteps & 62,250 \\
        Total RGB frames & 124,500 \\
        Frames per demo & 75--197 \\
        Mean frames per demo & 124.5 \\
        Cameras & 2 \\
        RGB resolution & $128 \times 128$ \\
        Canonical objects & 7 \\
        Directed relations & 8 \\
        Mean triplets / frame (\texttt{agentview}) & 42.0 \\
        Mean triplets / frame (\texttt{eye\_in\_hand}) & 16.73 \\
        \bottomrule
    \end{tabular}
\end{table}

The difference in graph density between the two views is intentional. The fixed \texttt{agentview} camera captures the global scene layout and therefore supports dense pairwise relation labels over the full object vocabulary. The wrist-mounted \texttt{eye\_in\_hand} camera observes a narrower and changing field of view, so its graph is filtered to visible objects. This paired-view structure allows EmbodimentSemantic to evaluate both global spatial ordering and egocentric spatial grounding.

\subsection{Real-robot dataset}
\begin{figure}[t]
    \centering
    \includegraphics[width=1\linewidth]{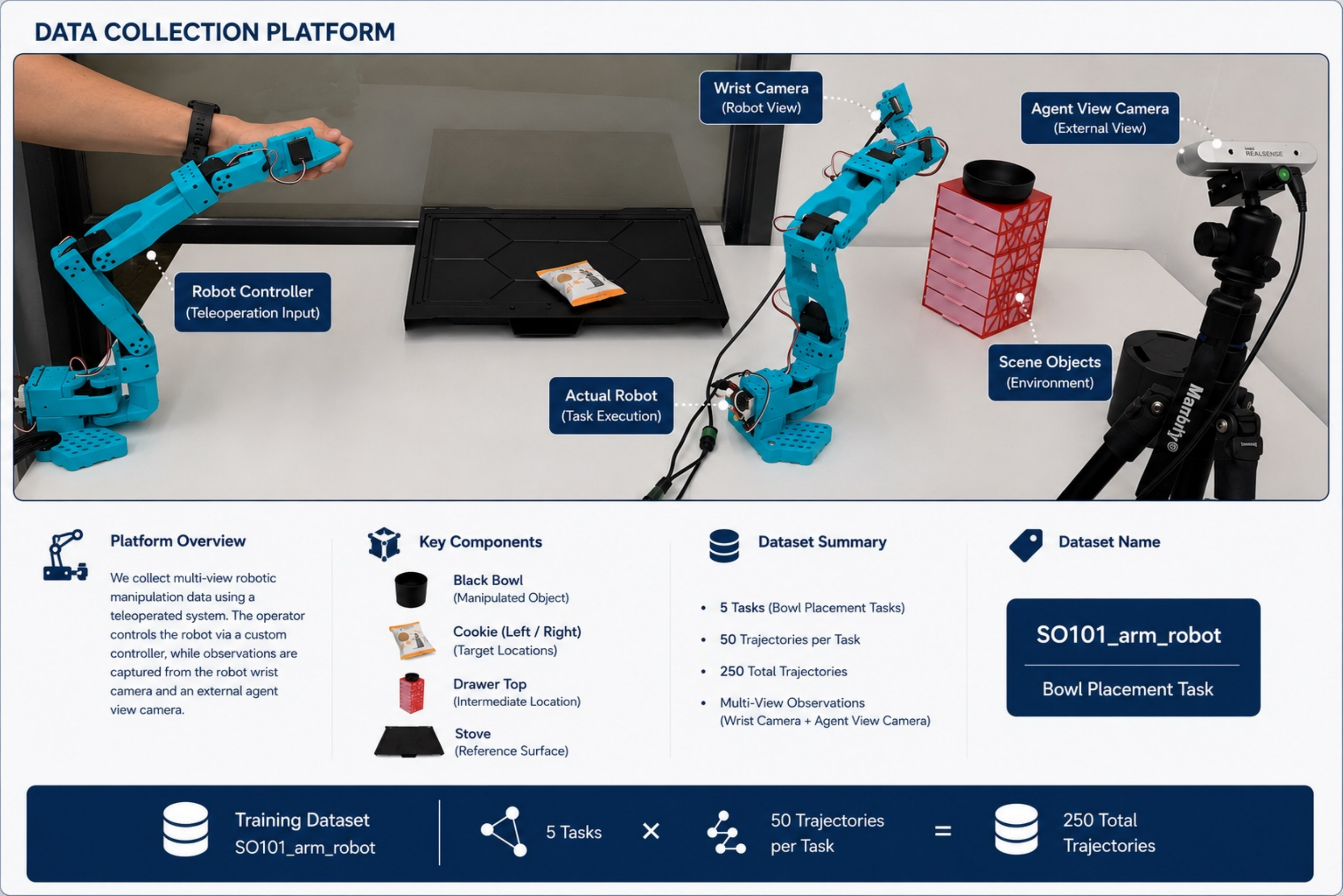}
    \caption{Teleoperation interface used to collect real-robot demonstrations for the bowl-placement tasks. The scene contains the SO101 robot arm, the manipulated black bowl, and task-relevant support and distractor objects.}
    \label{fig:teleoperation-setup}
\end{figure}
In addition to the simulator-grounded LIBERO benchmark, \textsc{EmbodimentSemantic} includes a real-robot bowl-placement dataset collected with a SO101 manipulation setup. The dataset  contains 257 teleoperated episodes across five tabletop manipulation tasks. Each task requires the robot to move a black bowl from a specified initial spatial context to a target support object, such as the stove or a cookie box.

The five real-robot tasks cover different starting relations for the manipulated bowl: on top of the drawer to the stove, left of the stove to the stove, right of the stove to the stove, on top of the drawer to the cookie box on the left, and on top of the drawer to the cookie box on the right. The demonstrations follow the LeRobot dataset format, with task-level metadata, trajectory data, and synchronized videos. The recorded video modalities include an external agent-view camera, a wrist camera, and an agent-view depth stream. Figure~\ref{fig:teleoperation-setup} shows the physical setup and teleoperation workflow used during data collection.
An example of the generated scene graphs are shown in Figure~\ref{fig:so101-scene-graph-examples}.

\section{Simulator-grounded graph generation}

EmbodimentSemantic generates frame-level scene graphs deterministically from MuJoCo simulator state. Rather than relying on manual relation annotations, the pipeline extracts object geometry, world coordinates, and camera projections for each demonstration frame. Offline annotations are generated for both \texttt{agentview} and \texttt{robot0\_eye\_in\_hand}. The online VLA context generator uses the same relation-assignment logic, but currently constructs live semantic context from the \texttt{agentview} camera.

\subsection{Object discovery, world coordinates, and bounding boxes}

For each demonstration frame, the system identifies task-relevant objects from MuJoCo bodies and excludes robot and gripper components. For every object, EmbodimentSemantic extracts the world-space position and projects its geometry into camera views. Offline annotation uses both the fixed third-person \texttt{agentview} camera and the wrist-mounted \texttt{robot0\_eye\_in\_hand} camera.

For each MuJoCo geometry primitive with world center \(p \in \mathbb{R}^3\), rotation matrix \(R\), and half-sizes \(s = (s_x, s_y, s_z)\), the eight local geometry corners are:
\begin{equation}
c(\sigma_x,\sigma_y,\sigma_z) =
\begin{bmatrix}
\sigma_x s_x \\
\sigma_y s_y \\
\sigma_z s_z
\end{bmatrix},
\qquad \sigma_x,\sigma_y,\sigma_z \in \{-1,+1\}.
\end{equation}
Each corner is transformed into world coordinates as:
\begin{equation}
x_w = p + R c.
\end{equation}
Given camera intrinsics \(K\) and inverse extrinsics \(E^{-1}\), world points are transformed into the camera frame:
\begin{equation}
\tilde{x}_c = E^{-1}
\begin{bmatrix}
x_w \\
1
\end{bmatrix}
=
\begin{bmatrix}
x_c \\
y_c \\
z_c \\
1
\end{bmatrix}.
\end{equation}
Points with \(z_c \leq 0.01\) are discarded, and the remaining points are projected using the standard pinhole camera model. The projected corners define an axis-aligned 2D bounding box, which is padded by 5 pixels and clamped to image bounds. Boxes that are implausibly large relative to the image size are discarded as projection artifacts.

This process produces, for each detected object, a world-space position and camera-specific projected bounding boxes. The world coordinates are used to assign spatial relations, while the projected boxes determine object visibility and support overlap-based vertical or containment checks.

\subsection{Spatial relation assignment}

Given a set of objects in a frame, EmbodimentSemantic constructs a directed scene graph by iterating over all ordered object pairs \((A,B)\), where \(A \neq B\). Each ordered pair receives one relation from the fixed ontology. Relation assignment follows two stages: an overlap-based vertical or containment check, followed by dominant-axis world-frame ordering.

First, EmbodimentSemantic checks whether the projected 2D boxes of \(A\) and \(B\) overlap strongly. Let the intersection rectangle be defined by \((\hat{x}_1,\hat{y}_1,\hat{x}_2,\hat{y}_2)\). We compute minimum-area overlap:
\begin{equation}
\mathrm{io}_{\min}(A,B) =
\frac{(\hat{x}_2-\hat{x}_1)(\hat{y}_2-\hat{y}_1)}
{\min\!\bigl((x_2^A-x_1^A)(y_2^A-y_1^A),\ (x_2^B-x_1^B)(y_2^B-y_1^B)\bigr)}.
\end{equation}
If \(\mathrm{io}_{\min}(A,B) > 0.8\) and at least one object is a bowl, the relation is assigned using the objects' world \(z\)-coordinates. For ordinary support cases, the higher object is labeled \texttt{is\_on\_top\_of} and the lower object is labeled \texttt{is\_below\_of}. For the drawer task, we use a task-specific containment rule. In Task 3, if \texttt{akita\_black\_bowl\_1} and \texttt{wooden\_cabinet\_1} satisfy the overlap condition, the pair is assigned \texttt{is\_inside} or \texttt{contains}. This hardcoded exception simplifies containment labeling for the only drawer-based task in EmbodimentSemantic.

If the vertical or containment condition does not apply, EmbodimentSemantic assigns a horizontal relation using world-frame object positions. Let:
\begin{equation}
\Delta x = x_A - x_B, \qquad \Delta y = y_A - y_B.
\end{equation}
If \(|\Delta x| \geq |\Delta y|\), the pair receives a depth relation, either \texttt{is\_in\_front\_of} or \texttt{is\_behind}. Otherwise, the pair receives a lateral relation, either \texttt{is\_left\_of} or \texttt{is\_right\_of}. The sign of the dominant offset determines the direction.

Importantly, EmbodimentSemantic defines left/right and front/behind relations in the MuJoCo world frame rather than directly in image-plane coordinates. This choice makes the labels consistent with the physical manipulation scene and simulator state, while camera projections determine which objects are visible in each view.

\subsection{Camera-specific graph construction}

EmbodimentSemantic augments the original LIBERO Spatial HDF5 files with additional semantic annotations; it does not regenerate or alter the original RGB observations. The stored RGB frames and projected bounding-box coordinates remain in the original LIBERO camera orientation. In particular, the released bounding boxes are defined with respect to the images as stored in the original HDF5 files.

Offline EmbodimentSemantic annotations are generated for both the fixed third-person \texttt{agentview} camera and the wrist-mounted \texttt{robot0\_eye\_in\_hand} camera. Spatial directions such as \texttt{is\_in\_front\_of}, \texttt{is\_behind}, \texttt{is\_left\_of}, and \texttt{is\_right\_of} are assigned from the signs and dominant axes of MuJoCo world-coordinate differences, not from image-plane pixel coordinates. This makes direction assignment deterministic rather than random, although the semantic naming of front/behind and left/right depends on the chosen world-frame convention.

The graph-generation rule is shared across the two camera views. For the wrist view, objects are first visibility-filtered using projected \texttt{robot0\_eye\_in\_hand} bounding boxes. In the current offline generation pipeline, relation assignment for the wrist graph then reuses the same world-frame relation logic used for \texttt{agentview}; wrist bounding boxes determine which objects are included, while MuJoCo world coordinates determine their spatial relations. Thus, the wrist graph is a visibility-filtered relational graph rather than a fully wrist-frame relation graph.

For visualization and human inspection, the \texttt{agentview} image may be rotated by \(180^\circ\) to match a more natural viewing orientation. This rotation is not applied to the stored LIBERO images or to the released bounding-box coordinates. The unrotated convention is preserved because downstream LIBERO VLA models are typically trained and evaluated on the original camera images as stored.

The resulting annotations are stored under camera-specific keys:
\begin{itemize}
    \item \texttt{agentview\_scene\_graph} for the fixed third-person camera;
    \item \texttt{robot0\_eye\_in\_hand\_scene\_graph} for the wrist-mounted camera.
\end{itemize}
Each graph is represented as a list of ordered triplets:
\[
(\texttt{objectA}, \texttt{relation}, \texttt{objectB}).
\]

\section{Experiments and results}

Our experiments evaluate whether VLMs can recover object-centric spatial scene graphs from embodied manipulation observations. Given an RGB robot observation, the model predicts directed object--relation--object triplets over a fixed set of objects and relations. We then evaluate whether each predicted relation is assigned to the correct ordered object pair.

This experiment is carried out in both simulated and real-world settings. In LIBERO-Spatial, the simulator state gives us ground-truth scene graphs for each selected frame, with observations from the simulated \texttt{agentview} and \texttt{wrist} cameras. The real-world SO101 dataset uses a matching external-camera and robot-camera setup with similar tabletop objects, allowing us to test the same scene-graph formulation on physical robot observations.

For each VLM, we use the same prompt and parsing protocol. The model receives the RGB observation, the task description, the object vocabulary, and the fixed set of relations and is asked to output one spatial triplet per line. We keep only valid triplets and compare the predictions against the reference graph for the corresponding task, episode, frame, and camera. Because the two black bowl instances are visually similar, evaluation treats predictions as invariant to swapping \texttt{akita\_black\_bowl\_1} and \texttt{akita\_black\_bowl\_2}. For each frame, we compute exact triplet F1 under the original prediction and under a swapped-bowl assignment, and score the higher-F1 assignment. This prevents penalizing models for visually ambiguous bowl-instance identity while still requiring exact relation and object-pair grounding for all other objects. We use mean per-task F1 as the main score for model comparison, since it gives equal weight to each manipulation task. In addition, we compute F1 in two complementary forms. Macro F1 first computes F1 independently for each evaluated frame and then averages these values, giving each frame equal weight. Micro F1 pools triplet-level true positives, false positives, and false negatives across all frames before computing the score. We report additional metrics including coverage, hallucination rate, reversal rate, direction consistency, per-relation performance, and per-object recall.

The rest of this section is organized as follows. We first report simulator-grounded VLM results on LIBERO, where ground-truth scene graphs allow exact quantitative evaluation. Within this controlled setting, the experiment is organized around three questions:
\begin{itemize}
    \item \textbf{Q1: Can VLMs recover exact spatial scene graphs from robot observations?} We answer this using simulator-grounded LIBERO frames with exact object--relation--object triplet evaluation.
    \item \textbf{Q2: Does viewpoint degrade spatial grounding?} We compare predictions from the fixed third-person \texttt{agentview} camera and the wrist-mounted \texttt{eye\_in\_hand} camera.
    \item \textbf{Q3: Which relation types are systematically hardest?} We analyze relation-type coverage and per-relation F1 to identify object-binding, depth, support, and containment failures.
\end{itemize}
We then discuss the real-world SO101 experiment and the released dataset, which includes real-world trajectories and generated scene-graph annotations.

\subsection{VLM benchmarking}

\paragraph{Simulator-grounded evaluation.}
We first evaluate VLMs on the simulator-grounded LIBERO benchmark. The controlled setting allows us to compare predicted scene graphs against reference graphs derived from simulator geometry. 
\begin{table}[t]
    \caption{Simulator-grounded VLM scene-graph prediction results. Mean per-task F1 is the main score. Macro F1 and micro F1 are computed as complementary aggregate scores but are not shown here for compactness.}
    \label{tab:vlm-sim-results}
    \centering
    \begin{tabular}{lrr}
        \toprule
        Model & \texttt{agentview}  mF1 & \texttt{eye\_in\_hand} mF1 \\
        \midrule
        \texttt{gemini-3.1-pro} & \textbf{0.5674} & \textbf{0.3773} \\
        \texttt{InternVL3-78B} & \underline{0.3519} & 0.2101 \\
        \texttt{Qwen3-VL-8B-Instruct} & 0.2837 & \underline{0.2108} \\
        \texttt{InternVL3\_5-14B} & 0.2561 & 0.1265 \\
        \texttt{gemma-4-E4B-it} & 0.2209 & 0.1472 \\
        \texttt{Molmo2-8B} & 0.1888 & 0.1459 \\
        \texttt{InternVL3\_5-8B} & 0.1769 & 0.1465 \\
        \texttt{nemotron-nano-12b-v2-vl} & 0.1599 & 0.1270 \\
        \texttt{Qwen2.5-VL-7B-Instruct} & 0.1405 & 0.1147 \\
        \texttt{MomaGraph-R1} & 0.1023 & 0.0926 \\
        \bottomrule
    \end{tabular}
\end{table}

The results show that exact spatial scene-graph prediction remains difficult across models. The proprietary \texttt{gemini-3.1-pro} model achieves the best performance on both views, with a mean per-task F1 of 0.5674 on \texttt{agentview} and 0.3773 on \texttt{eye\_in\_hand}. This sets the strongest reference point for the benchmark. Among open-source models, \texttt{InternVL3-78B} performs best on \texttt{agentview}, while \texttt{Qwen3-VL-8B-Instruct} performs best on \texttt{eye\_in\_hand}. Performance is lower on the wrist camera for most models, which is expected because the view is narrower, changes with robot motion, and contains partial observations of the fixed object set from which relations must be inferred.

The gap between the two camera views is important. The \texttt{agentview} camera provides a stable view of the full workspace with one object being moved. This tests whether VLMs update their predicted relations as the scene changes over time. While \texttt{eye\_in\_hand} requires the model to reason from an egocentric and changing perspective. This makes the wrist-camera setting a stricter test of viewpoint-dependent spatial grounding. The lower wrist-view scores suggest that even the strongest open-source models still struggle when spatial layout must be inferred from partial and moving robot observations.

Mean per-task F1 gives the main comparison, while the broader metric set exposes related failure modes. Macro and micro F1 provide frame-level and pooled triplet-level views of the same exact-matching problem. Relation-type coverage measures the fraction of reference triplets whose predicate type appears anywhere in the model’s prediction for the same frame, while hallucination rate measures over-generation. Reversal rate and direction consistency test whether the model preserves the ordered structure of relations. These diagnostics are useful because a model can predict plausible spatial predicates while still assigning them to the wrong object pair or reversing their direction.

\begin{table}[t]
\centering
\small
\caption{Main VLM diagnostics on \texttt{agentview}. Coverage is relation-type coverage, while F1 requires exact object--relation--object triplet matching. The large coverage--F1 gap shows that models predict plausible predicates but often bind them to the wrong ordered object pairs. Per-relation F1 is averaged across evaluated VLMs.}
\label{tab:vlm-diagnostics}
\begin{tabular}{lcc}
\toprule
Diagnostic & Value & Interpretation \\
\midrule
Best coverage & 0.980 & Relation types are often mentioned \\
Best exact mF1 & 0.567 & Exact triplets remain difficult \\
Gemini coverage / F1 & 0.968 / 0.567 & Strongest model still has binding errors \\
Gemma coverage / F1 & 0.969 / 0.221 & High coverage, low exact grounding \\
MomaGraph coverage / F1 & 0.947 / 0.102 & Plausible predicates, poor binding \\
\midrule
\texttt{is\_left\_of} F1 & 0.278 & Easiest relation family \\
\texttt{is\_right\_of} F1 & 0.276 & Easiest relation family \\
\texttt{is\_in\_front\_of} F1 & 0.233 & Depth remains difficult \\
\texttt{is\_behind} F1 & 0.222 & Depth remains difficult \\
\texttt{is\_on\_top\_of} F1 & 0.196 & Support relation is harder \\
\texttt{is\_inside} F1 & 0.191 & Containment is harder \\
\texttt{is\_below\_of} F1 & 0.189 & Vertical inverse is harder \\
\texttt{contains} F1 & 0.153 & Hardest relation \\
\bottomrule
\end{tabular}
\end{table}
Although several models achieve high relation-type coverage, their exact triplet F1 remains much lower, revealing a core failure mode that aggregate VQA-style spatial benchmarks can hide. For example, \texttt{gemini-3.1-pro} reaches 0.968 coverage on \texttt{agentview}, but only 0.567 mean per-task F1; similarly, \texttt{gemma-4-E4B-it} reaches 0.969 coverage but only 0.221 F1, and \texttt{MomaGraph-R1} reaches 0.947 coverage but only 0.102 F1. This gap shows that current VLMs often recover plausible spatial predicates while failing to bind them to the correct ordered object pairs. Per-relation results in Table~\ref{tab:vlm-diagnostics} further show that failures are systematic: lateral relations are easier than depth, support, and containment relations, with \texttt{contains} performing worst overall. Thus, the dominant failure is not spatial vocabulary, but exact object-pair grounding under directed scene-graph evaluation.

The diagnostic metrics show that these errors are structured rather than random. Across complete \texttt{agentview} runs, the average hallucination rate is 0.690, indicating that VLMs frequently over-generate plausible but incorrect spatial triplets; at the same time, average direction consistency is only 0.205, showing that models rarely recover both directions of inverse relation pairs correctly. Reversal errors further indicate failures of asymmetric grounding: \texttt{qwen3-vl-8b} has the highest reversal rate at 0.288, while \texttt{InternVL3-78B}, \texttt{InternVL3\_5-14B}, and \texttt{gemma-4-E4B-it} also show substantial reversal rates of 0.163, 0.149, and 0.133. Together with the relation-level gradient in Table~\ref{tab:vlm-diagnostics}, these results suggest that current VLMs often detect spatial co-occurrence and produce reasonable relation words, but struggle to resolve ordered, asymmetric spatial assignments from RGB observations, especially for depth, support, and containment relations.

\paragraph{Real-world evaluation.}
We also apply the same scene-graph prompting framework to the real-world SO101 dataset. This split is not used for quantitative scoring, since it does not currently include reference scene-graph annotations. Instead, we use the best-performing VLM from the previous experiment, \texttt{gemini-3.1-pro}, to generate scene graphs for physical robot observations.

The SO101 tasks are designed to mirror the spatial structure of LIBERO Spatial with similar tabletop objects and external and wrist camera views. The model is prompted with the same relation ontology and a corresponding object vocabulary, producing directed object--relation--object triplets for sampled frames. These predictions are included as a real-world scene-graph resource rather than as ground truth.

We release both components of \textsc{EmbodimentSemantic}: the real-world SO101 manipulation dataset and the simulator-grounded LIBERO annotations. The release includes physical robot observations, generated real-world scene graphs, and more than 120K camera-specific simulator-grounded scene graphs. Together, these components connect controlled simulator labels with real robot data and provide a resource for studying spatial grounding in embodied manipulation.

\subsection{VLA Benchmarking}

EmbodimentSemantic also provides an online VLA evaluation interface for testing how explicit spatial context affects policy behavior in LIBERO-Spatial. Unlike the offline VLM benchmark, where scene graphs are precomputed and used as ground truth for triplet prediction, the VLA interface generates scene graphs during policy evaluation and injects them into the task prompt consumed by existing fine-tuned VLA policies.

The interface wraps the LeRobot evaluation environment for LeRobot-supported policies, including Pi0 and Pi05. It augments each task description with semantic context computed from the current simulator state. At evaluation time, the system extracts object geometry, projected bounding boxes, and world-frame spatial relations from MuJoCo, serializes this information as text, and appends it to the original task instruction. This allows EmbodimentSemantic to evaluate existing policies under different spatial-context conditions without changing the policy architecture or retraining the model.

\paragraph{Real-time semantic context generation}

Live scene graphs are generated from the current simulator state using the same world-frame relation assignment logic as the offline annotation pipeline. In the current interface, live semantic context is generated from the \texttt{agentview} camera. The injected graph can also be subject-filtered. In our default configuration, the graph is centered on \texttt{akita\_black\_bowl\_1}, the manipulated object in all EmbodimentSemantic tasks. This keeps the prompt focused on task-relevant spatial relations rather than injecting the full dense graph. Removing the filter exposes the full graph, enabling controlled comparison between focused and exhaustive relational context.

\paragraph{Task-conditioned scene randomization}

Randomization is a core part of the online VLA evaluation interface. Standard LIBERO evaluation uses fixed task templates and familiar object layouts; EmbodimentSemantic adds controlled perturbations to test whether policies remain robust when the spatial arrangement changes. The goal is not to create arbitrary random scenes, but to generate reproducible perturbations that preserve the task goal while changing the spatial context around the manipulated object.

At environment reset, the randomization module can apply task-specific perturbations such as object pose swaps, relative object moves, and fixed pose offsets. After perturbing the scene, the simulator is stepped to let moved objects settle while preserving the robot's initial configuration. This produces randomized but valid evaluation scenes for testing whether policies depend on memorized LIBERO layouts or respond to the current spatial configuration.

The randomization rules are task-conditioned by design. Some perturbations would invalidate the task, remove a required support object, or break the initial placement sampler. EmbodimentSemantic therefore uses explicit per-task perturbation rules so that robustness evaluations remain meaningful and reproducible.

\paragraph{Object, prompt, and camera perturbations}

In addition to pose-level randomization, the interface supports object, prompt, and camera perturbations.

First, selected distractor objects can be removed before environment construction. The removal procedure preserves task validity by avoiding objects that appear in the goal condition or are required for the manipulated bowl's initial placement. This allows the benchmark to test whether policies rely on irrelevant distractors or on the true task-relevant spatial structure.

Second, selected task prompts can be rewritten while preserving the underlying goal condition. Since LIBERO success is determined by the environment goal rather than the exact prompt string, prompt overrides allow controlled analysis of how policies respond to different linguistic descriptions of the same task.

Third, the camera configuration can be changed for selected tasks. This enables viewpoint-shift evaluation, including camera views outside the default training setup. Such changes intentionally test out-of-distribution visual conditions for policies trained primarily on the standard LIBERO cameras.
Table~\ref{tab:vla-randomizations} summarizes the task-conditioned perturbations used in the online VLA evaluation interface.

\begin{table}[t]
    \caption{Task-conditioned perturbations implemented in the EmbodimentSemantic online VLA evaluation interface. Perturbations are applied to selected LIBERO-Spatial tasks while preserving the bowl-to-plate goal when possible.}
    \label{tab:vla-randomizations}
    \centering
    \scriptsize
    \begin{tabularx}{\linewidth}{l l X}
        \toprule
        Perturbation type & Task IDs & Implemented perturbation \\
        \midrule
        Subject-filtered graph injection 
        & All tasks 
        & Inject only scene-graph triplets whose subject is \texttt{akita\_black\_bowl\_1}; disabling the filter injects the full scene graph. \\
        \midrule
        Object removal 
        & T4 
        & Remove \texttt{cookies\_1} and \texttt{glazed\_rim\_porcelain\_ramekin\_1} from the drawer task to clear distractors from the table. \\
        & T8 
        & Remove \texttt{glazed\_rim\_porcelain\_ramekin\_1} and \texttt{cookies\_1} from the next-to-plate task to remove decoys. \\
        \midrule
        Prompt override 
        & T0 
        & Replace the original prompt with ``pick up the black bowl in front of the ramekin and place it on the plate.'' \\
        & T7 
        & Replace the original prompt with ``pick up the black bowl behind the wooden cabinet and place it on the plate.'' \\
        \midrule
        Camera override 
        & T2 
        & Replace the default \texttt{agentview}/\texttt{robot0\_eye\_in\_hand} cameras with \texttt{frontview} and \texttt{robot0\_robotview}. \\
        & T6 
        & Replace the default \texttt{agentview}/\texttt{robot0\_eye\_in\_hand} cameras with \texttt{frontview} and \texttt{robot0\_robotview}. \\
        \midrule
        Pose swaps / moves 
        & T1 
        & Swap \texttt{akita\_black\_bowl\_1} with \texttt{akita\_black\_bowl\_2}; swap \texttt{glazed\_rim\_porcelain\_ramekin\_1} with \texttt{cookies\_1}; swap \texttt{cookies\_1} with \texttt{plate\_1}. \\
        & T3 
        & Swap \texttt{akita\_black\_bowl\_2} with \texttt{plate\_1}; swap \texttt{glazed\_rim\_porcelain\_ramekin\_1} with \texttt{akita\_black\_bowl\_2}. \\
        & T5 
        & Swap \texttt{cookies\_1} with \texttt{glazed\_rim\_porcelain\_ramekin\_1}; swap \texttt{akita\_black\_bowl\_1} with \texttt{akita\_black\_bowl\_2}; move \texttt{akita\_black\_bowl\_1} by \((0.0,0.0,0.05)\); move \texttt{plate\_1} by \((-0.05,-0.45,0.5)\). \\
        & T9 
        & Swap \texttt{akita\_black\_bowl\_2} with \texttt{plate\_1}; swap \texttt{cookies\_1} with \texttt{akita\_black\_bowl\_2}; move \texttt{akita\_black\_bowl\_2} by \((0.0,-0.1,0.0)\). \\
        \bottomrule
    \end{tabularx}
\end{table}

\paragraph{Evaluation modes and ablations}
The online VLA benchmark evaluates policies under two prompt conditions:
\begin{itemize}
    \item \texttt{standard}: the policy receives the standard LIBERO task prompt under the selected perturbation setting;
    \item \texttt{scene\_graph}: the policy receives the same LIBERO task prompt augmented with live EmbodimentSemantic scene-graph information.
\end{itemize}

We evaluate two policies fine-tuned on the LIBERO Spatial dataset: Pi0 (\texttt{lerobot/pi0\_libero\_finetuned}) and Pi05 (\texttt{lerobot/pi05\_libero\_finetuned}). Both are evaluated using the LeRobot evaluation environment.

The ablation compares the standard prompt against the scene-graph-augmented prompt under the same task and perturbation settings. This provides a matched comparison of the same policy weights, task goal, and scene perturbation under standard and scene-graph-augmented prompts.

A key risk in both VLM and VLA evaluation is prompt shortcutting. LIBERO task descriptions often contain spatial information, such as ``between the plate and the ramekin'' or ``in the top drawer.'' A model may exploit these language priors without grounding them in the current observation. EmbodimentSemantic therefore supports ablations that separate visual grounding, task-language priors, and misleading task context.

Together, real-time graph injection, subject-filtered graph context, and task-conditioned randomization make EmbodimentSemantic an action-level evaluation interface for spatial grounding. The interface tests not only whether a policy succeeds under the default LIBERO scene, but also how it responds when the spatial context, object layout, prompt condition, or viewpoint is systematically changed.

\paragraph{VLA Benchmarking Results}

Table~\ref{tab:vla-results} reports the per-task VLA success rates for the standard and scene-graph prompt conditions. For each policy and task condition, we report success over 50 evaluation episodes. In the table, \texttt{std} denotes the standard LIBERO prompt and \texttt{sg} denotes the same prompt augmented with live scene-graph context.

The results show that scene-graph context can improve policy success on some tasks, but the effect is policy- and task-dependent. For Pi0, scene-graph prompting improves Task 2 from 90.0\% to 94.0\% success. Task 6 remains unchanged at 90.0\%, Task 0 remains unchanged at 2.0\%, and the remaining Pi0 tasks remain at 0.0\% under both prompt conditions.

For Pi05, the largest gain occurs on Task 4, where success increases from 46.0\% under the standard prompt to 76.0\% with scene-graph context. Smaller gains appear on Task 5, from 0.0\% to 2.0\%, and Task 7, from 8.0\% to 16.0\%. Task 0 remains saturated at 100.0\% in both conditions.

The table also shows that scene-graph injection is not uniformly beneficial. On Task 8 for Pi05, success drops from 44.0\% to 24.0\%, suggesting that added relational text can interfere with policy behavior when the prompt becomes less aligned with the policy's learned task representation. Overall, these results show that \textsc{EmbodimentSemantic} provides a practical interface for diagnosing VLA policies: explicit spatial context can produce substantial gains on selected tasks while also revealing cases where stronger policy grounding or prompt conditioning is still needed.

\begin{table*}[t]
\centering
\small
\caption{
Task success (mean over 50 evaluation episodes). 
We report success rates for the baseline policy (\texttt{std}) and with injected scene graphs (\texttt{sg}) under two LIBERO-finetuned policies. 
$\Delta$ denotes the change in success rate (percentage points, pp) when using \texttt{sg} vs.~\texttt{std}.
}
\label{tab:vla-results}

\setlength{\tabcolsep}{6pt}
\renewcommand{\arraystretch}{1.15}

\begin{tabular}{c c c c c c c}
\toprule

\multirow{2}{*}{Task ID}
& \multicolumn{3}{c}{\texttt{pi0\_libero\_finetuned}}
& \multicolumn{3}{c}{\texttt{pi05\_libero\_finetuned}} \\

\cmidrule(lr){2-4}
\cmidrule(lr){5-7}

&
\texttt{std} (\%)
& \texttt{sg} (\%)
& $\Delta$ (pp)
& \texttt{std} (\%)
& \texttt{sg} (\%)
& $\Delta$ (pp) \\

\midrule

0
& 2.0 (1/50)
& 2.0 (1/50)
& +0.0
& 100.0 (50/50)
& 100.0 (50/50)
& +0.0 \\

1
& 0.0 (0/50)
& 0.0 (0/50)
& +0.0
& 0.0 (0/50)
& 0.0 (0/50)
& +0.0 \\

2
& 90.0 (45/50)
& \textbf{94.0 (47/50)}
& \textcolor{green!50!black}{+4.0}
& 0.0 (0/50)
& 0.0 (0/50)
& +0.0 \\

3
& 0.0 (0/50)
& 0.0 (0/50)
& +0.0
& 0.0 (0/50)
& 0.0 (0/50)
& +0.0 \\

4
& 0.0 (0/50)
& 0.0 (0/50)
& +0.0
& 46.0 (23/50)
& \textbf{76.0 (38/50)}
& \textcolor{green!50!black}{+30.0} \\

5
& 0.0 (0/50)
& 0.0 (0/50)
& +0.0
& 0.0 (0/50)
& \textbf{2.0 (1/50)}
& \textcolor{green!50!black}{+2.0} \\

6
& 90.0 (45/50)
& 90.0 (45/50)
& +0.0
& 0.0 (0/50)
& 0.0 (0/50)
& +0.0 \\

7
& 0.0 (0/50)
& 0.0 (0/50)
& +0.0
& 8.0 (4/50)
& \textbf{16.0 (8/50)}
& \textcolor{green!50!black}{+8.0} \\

8
& 0.0 (0/50)
& 0.0 (0/50)
& +0.0
& \textbf{44.0 (22/50)}
& 24.0 (12/50)
& \textcolor{red}{-20.0} \\

9
& 0.0 (0/50)
& 0.0 (0/50)
& +0.0
& 0.0 (0/50)
& 0.0 (0/50)
& +0.0 \\

\bottomrule
\end{tabular}

\vspace{0.5em}

\footnotesize
\textit{std}: baseline policy without scene graphs.
\textit{sg}: policy with injected scene graphs.
Positive $\Delta$ indicates improvement; negative $\Delta$ indicates degradation.

\end{table*}


\section{Limitations}

EmbodimentSemantic is designed as a controlled diagnostic benchmark, and its scope is intentionally narrow. The current dataset focuses on one LIBERO-Spatial task family, seven canonical objects, and a fixed eight-predicate relations set. This controlled setting makes exact triplet-level evaluation possible, but it limits claims about broader open-vocabulary spatial grounding or real-world generalization.

The benchmark also relies on simulator-derived labels. This avoids manual annotation noise and provides dense frame-level supervision, but the resulting relations depend on the chosen MuJoCo world-frame convention, visibility filtering rules, and task-specific containment heuristics. In particular, left/right and front/behind are defined by world-frame axes rather than image-plane coordinates. This makes the labels consistent with the manipulation scene, but requires care when interpreting camera-specific model predictions.

EmbodimentSemantic evaluates spatial relation prediction under a fixed object vocabulary. Object names are provided to the model, so the benchmark isolates relational grounding rather than open-vocabulary object detection. This is intentional, but it means that strong performance on EmbodimentSemantic would not by itself imply robust object recognition in unconstrained robot scenes.

Finally, the VLA interface evaluates how explicit spatial context can be injected into existing policies, but action-level conclusions require controlled policy runs under matched tasks, seeds, context modes, and perturbation settings. The current contribution is therefore best understood as a benchmark and evaluation interface for spatial grounding, not as a new VLA architecture.

\section{Conclusion and Future Work}

We introduced EmbodimentSemantic, a simulator-grounded benchmark for evaluating spatial scene-graph prediction in robot manipulation scenes. EmbodimentSemantic converts LIBERO Spatial demonstrations into frame-level object-relation-object triplets by extracting MuJoCo geometry, world coordinates, and projected bounding boxes from robot camera observations. This produces a controlled offline benchmark for VLM spatial grounding and an online interface for injecting real-time scene graphs into VLA policy evaluation.

Our VLM results show that current models can achieve high relation coverage while still obtaining low exact triplet F1, indicating that plausible spatial language does not necessarily imply precise relational grounding. By evaluating object pairs, relation direction, hallucination, and viewpoint-specific structure directly, EmbodimentSemantic exposes spatial grounding as a measurable bottleneck for embodied foundation models.

Future work will extend EmbodimentSemantic beyond the current LIBERO Spatial slice. A natural next step is to generate scene-graph annotations for the full LIBERO suite, including object, goal, long-horizon, and multi-task settings. This would increase object diversity, relation diversity, task complexity, and scene variation while preserving the simulator-grounded annotation pipeline.

A second direction is to use EmbodimentSemantic as training data rather than only as an evaluation benchmark. The generated triplets, bounding boxes, and world-coordinate annotations can supervise VLMs to produce structured scene graphs from robot observations. This would allow direct comparison between off-the-shelf VLM prompting, supervised fine-tuning, and multimodal instruction tuning for spatial graph prediction.

A third direction is to fine-tune VLA policies with EmbodimentSemantic context. Instead of injecting scene graphs only at evaluation time, future policies could be trained to condition on structured relational context, bounding boxes, or both. This would test whether explicit spatial representations improve action prediction, robustness to layout shifts, and generalization under changed viewpoints or object configurations.

Future work can also expand the relation ontology. The current benchmark focuses on eight directed spatial predicates, but manipulation often depends on richer relations such as contact, support, occlusion, reachability, containment depth, graspability, and task-relevant affordances. Extending EmbodimentSemantic from geometric relations to affordance-aware scene graphs would make the representation more directly useful for planning and policy learning.

Finally, EmbodimentSemantic opens the door to closed-loop spatial evaluation. Because the online interface can generate scene graphs during policy execution, future work can evaluate whether policies not only perceive spatial relations correctly, but also update their behavior when those relations change. This would move beyond static frame-level evaluation toward interactive tests of spatial grounding, correction, and recovery in embodied agents.

\clearpage

\bibliography{example}  

\appendix

\section{Appendix}
\label{app:vlm-evaluation-plots}

\begin{figure}[p]
    \centering
    \includegraphics[width=0.48\linewidth]{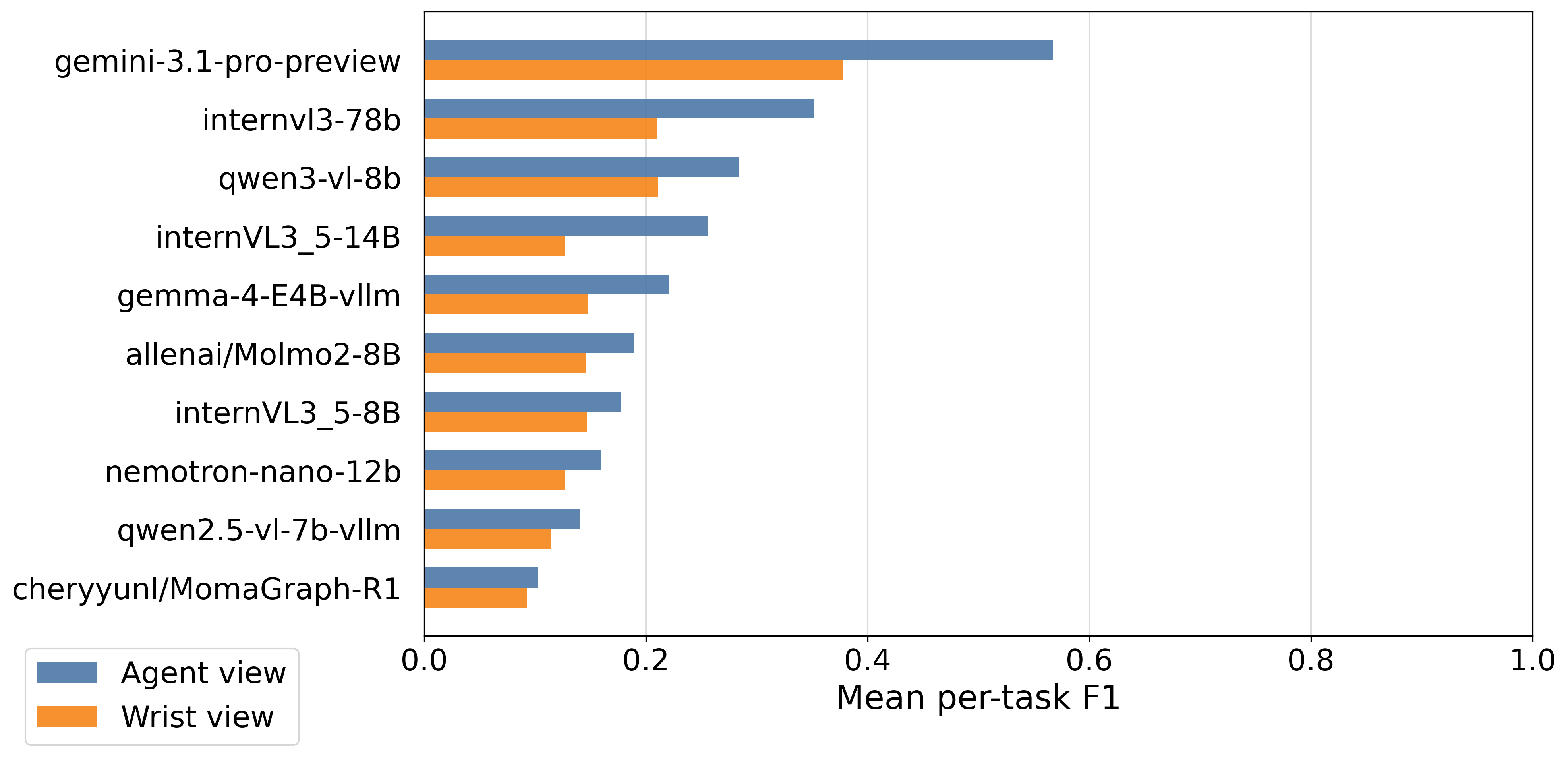}
    \includegraphics[width=0.48\linewidth]{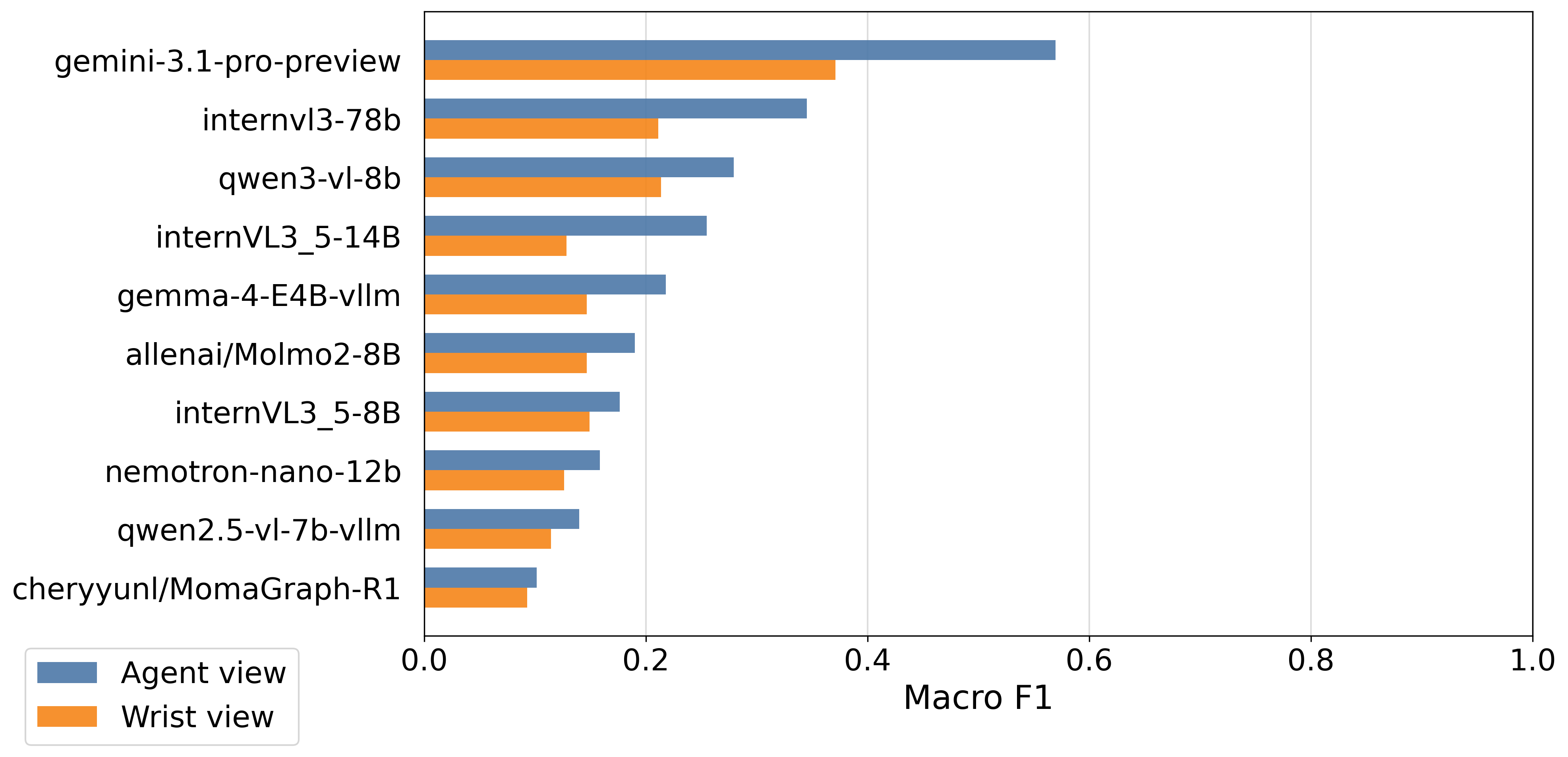}
    \includegraphics[width=0.48\linewidth]{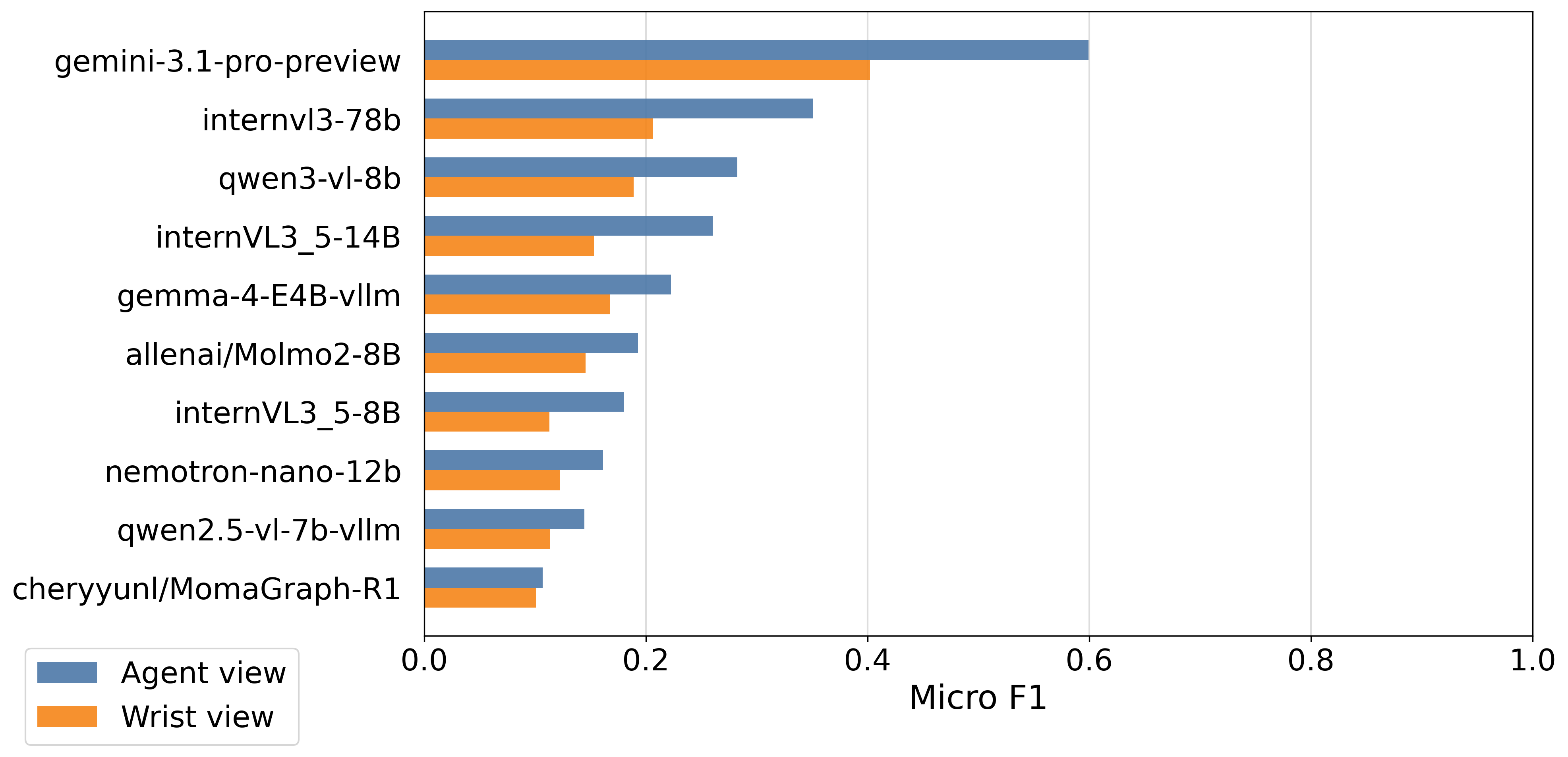}
    \includegraphics[width=0.48\linewidth]{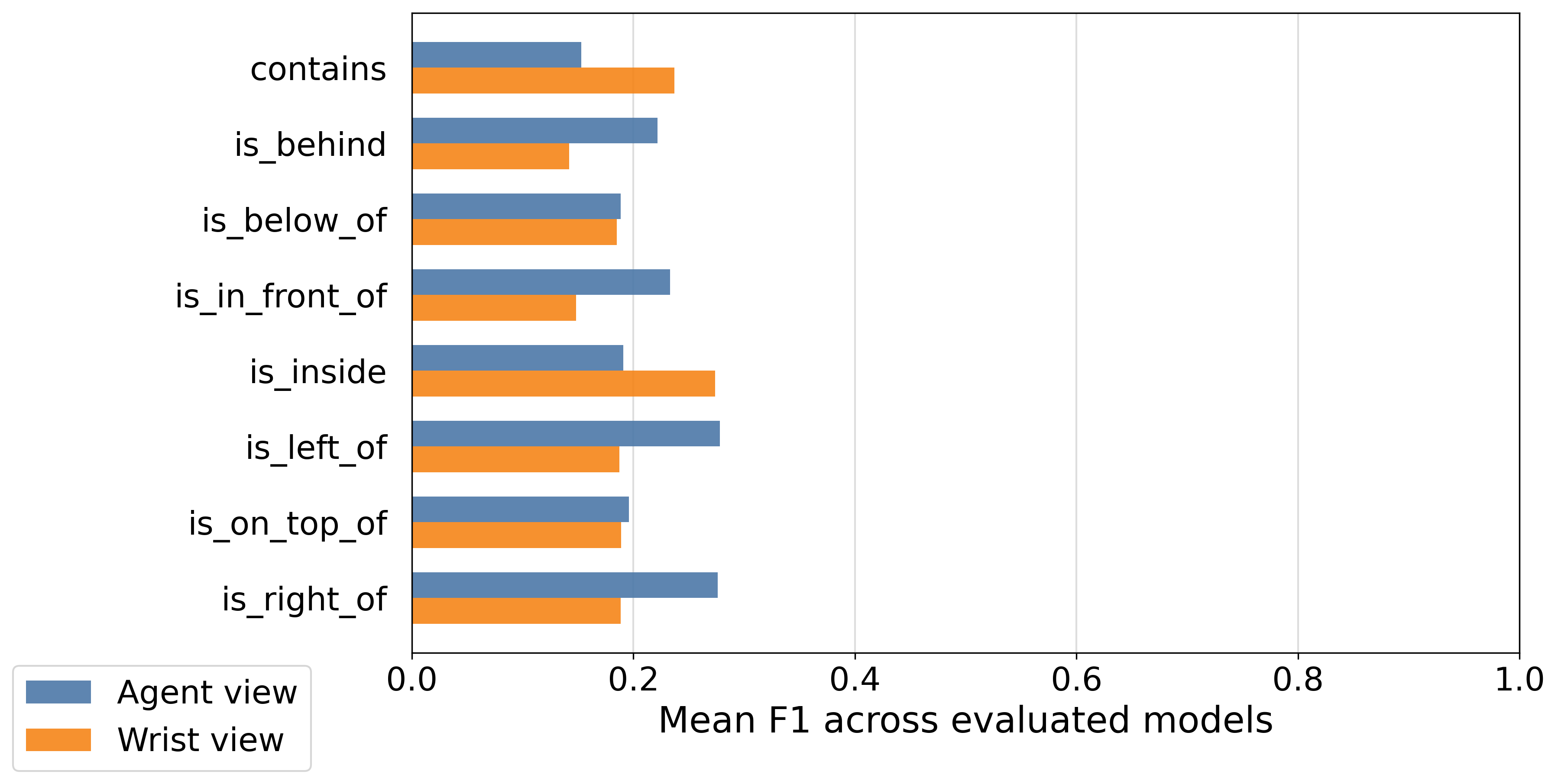}
    \caption{Additional aggregate F1 results for VLM scene-graph prediction.}
    \label{fig:app-vlm-f1}
\end{figure}

\begin{figure}[p]
    \centering
    \includegraphics[width=0.48\linewidth]{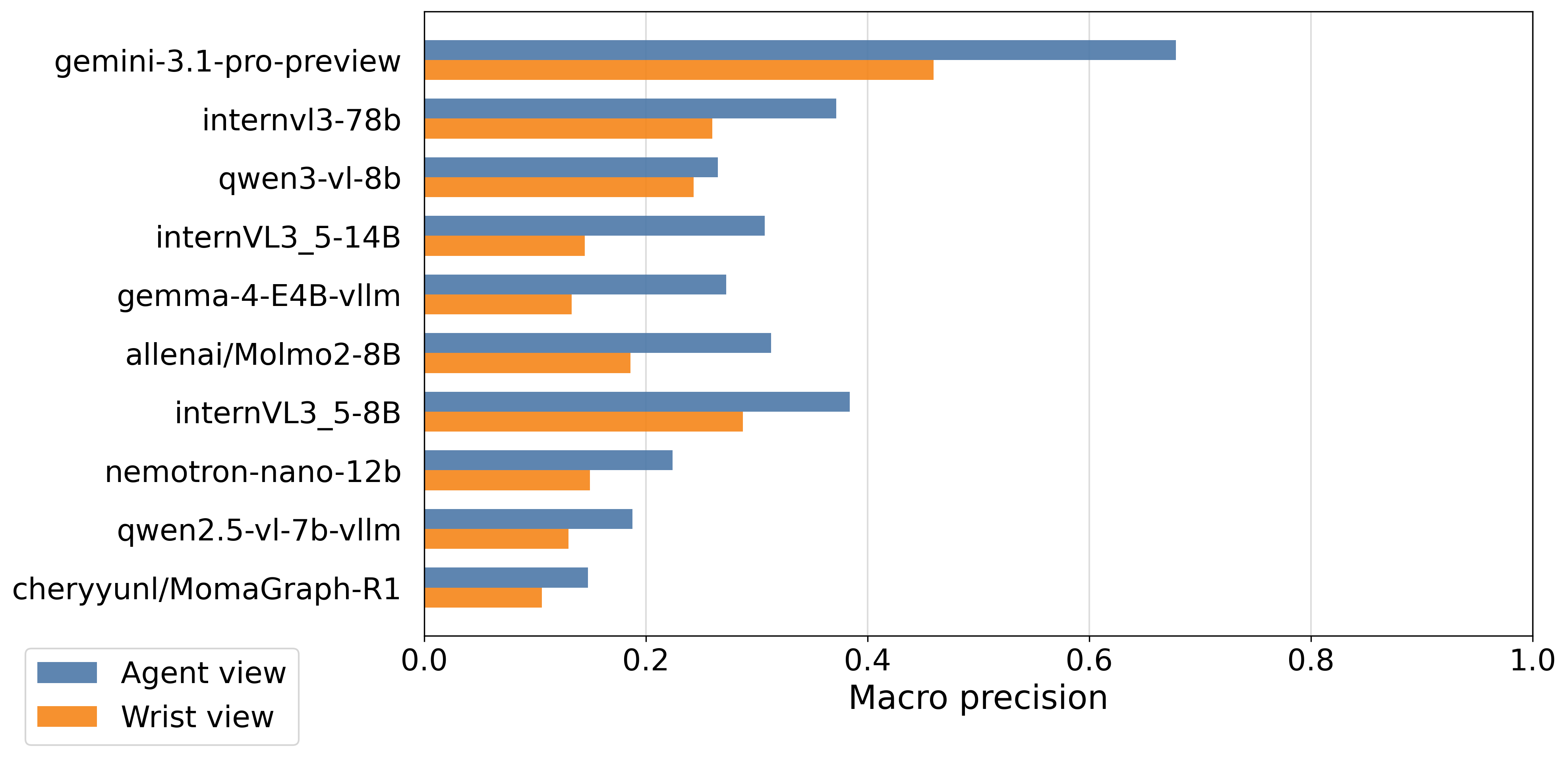}
    \includegraphics[width=0.48\linewidth]{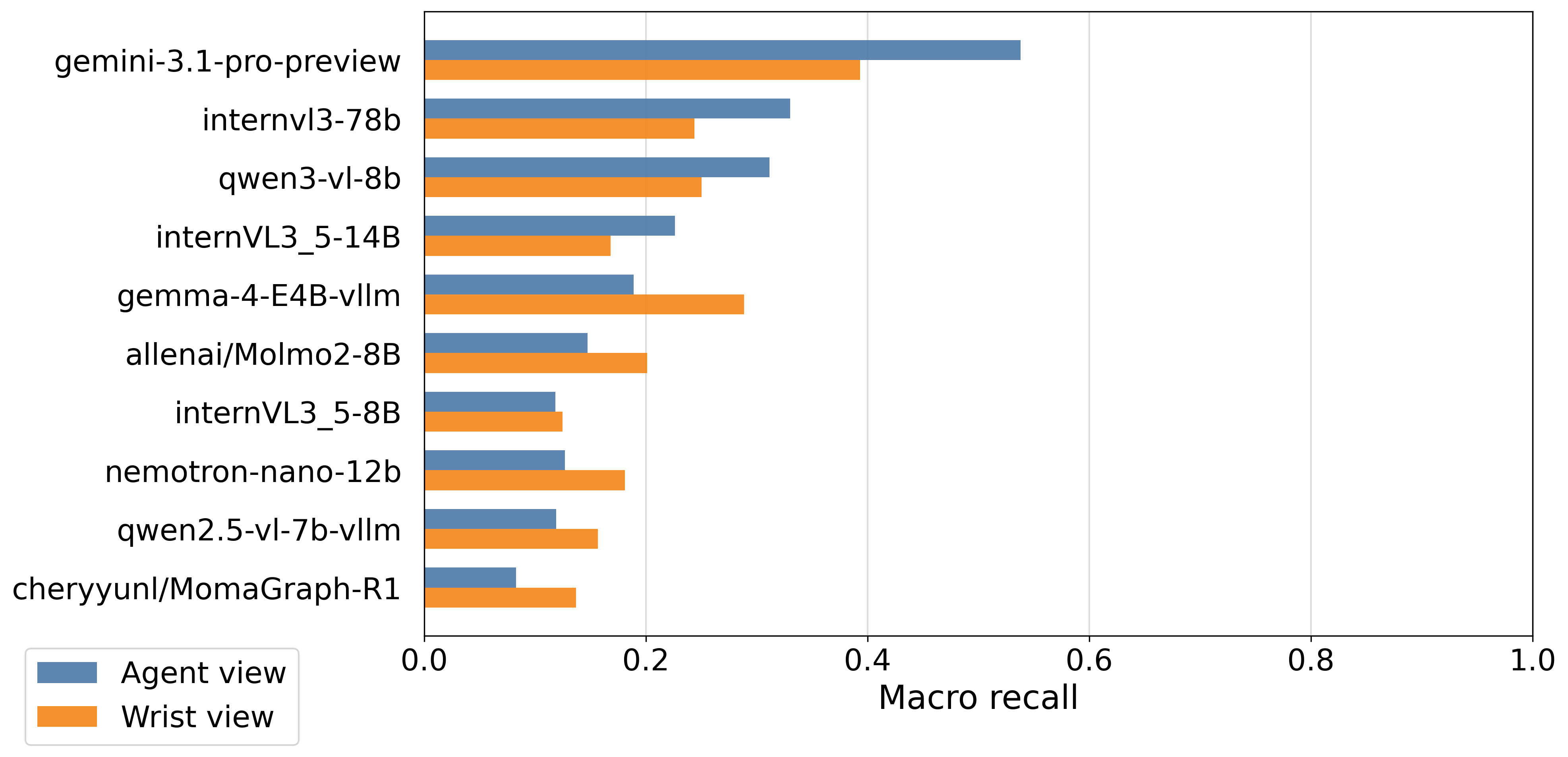}
    \includegraphics[width=0.48\linewidth]{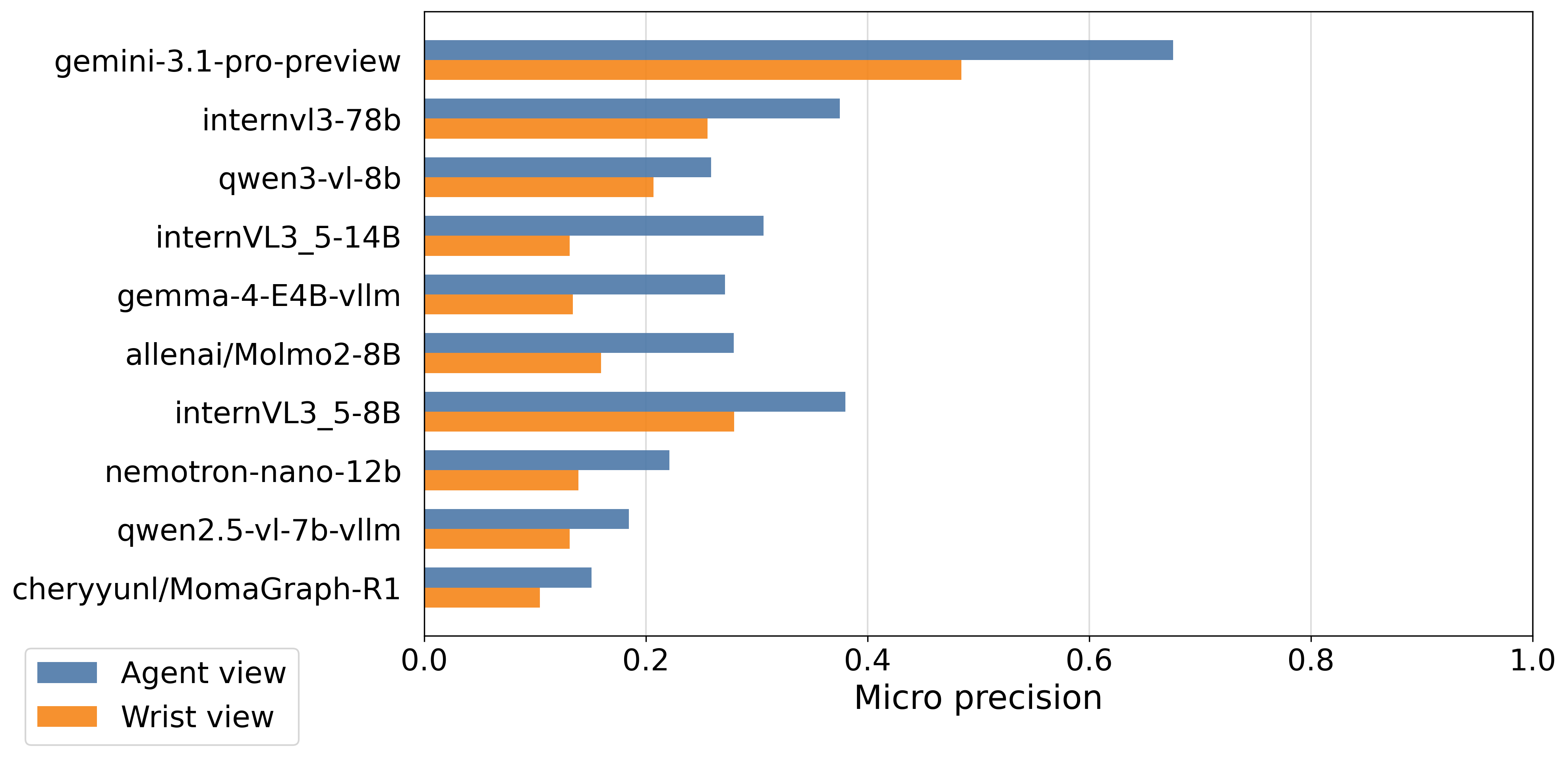}
    \includegraphics[width=0.48\linewidth]{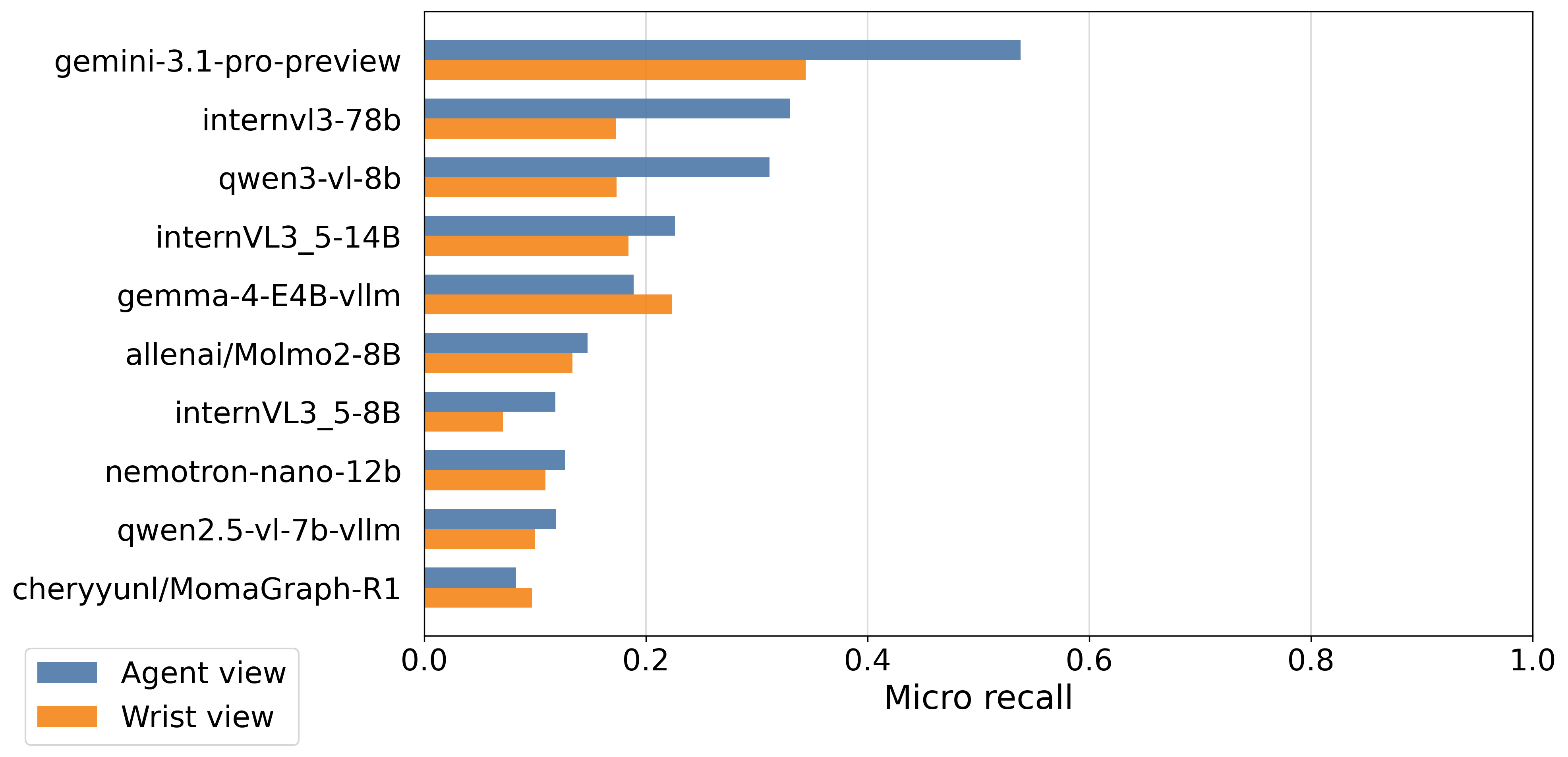}
    \caption{Additional precision and recall results for VLM scene-graph prediction.}
    \label{fig:app-vlm-pr}
\end{figure}

\begin{figure}[p]
    \centering
    \includegraphics[width=0.48\linewidth]{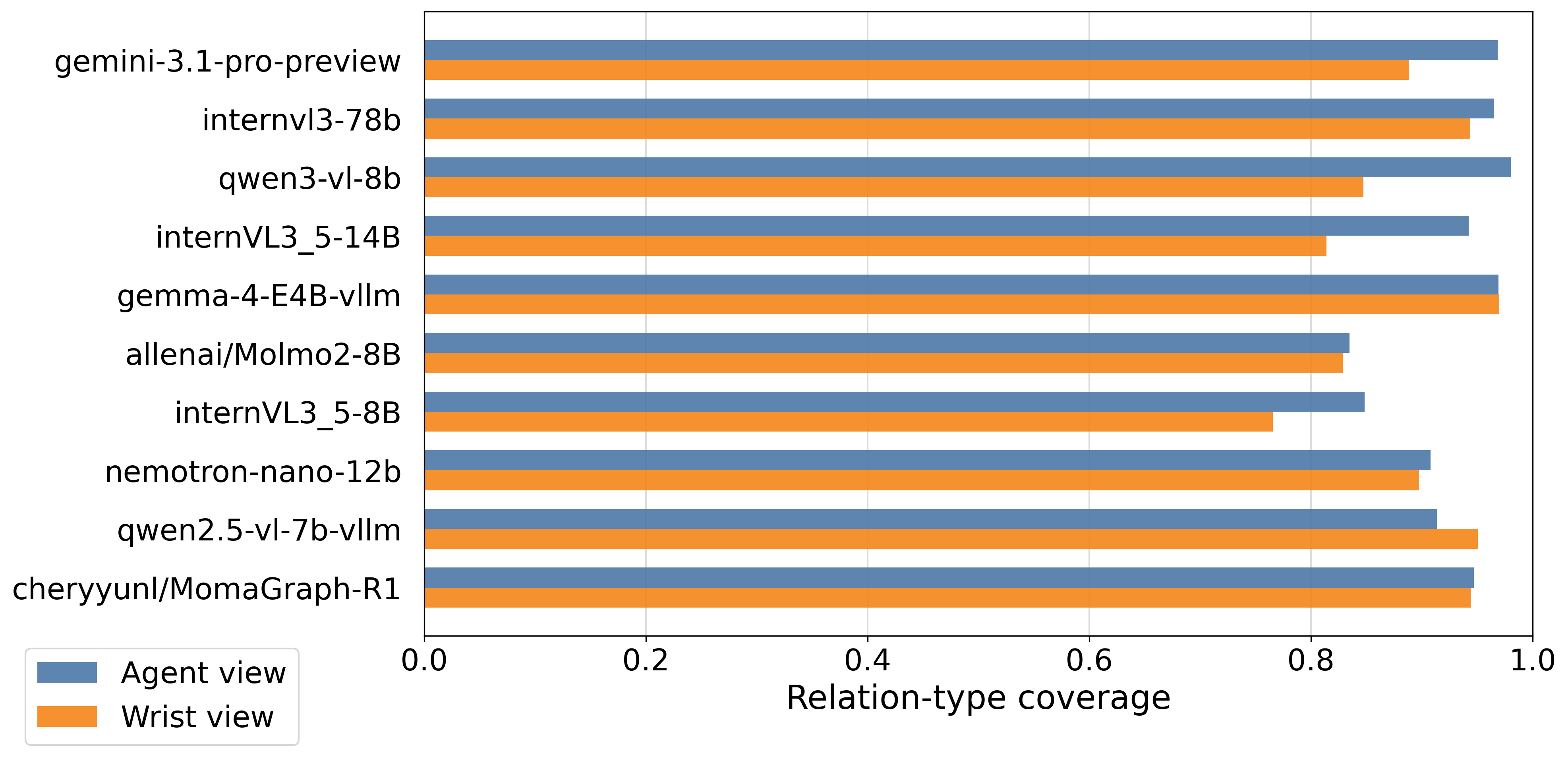}
    \includegraphics[width=0.48\linewidth]{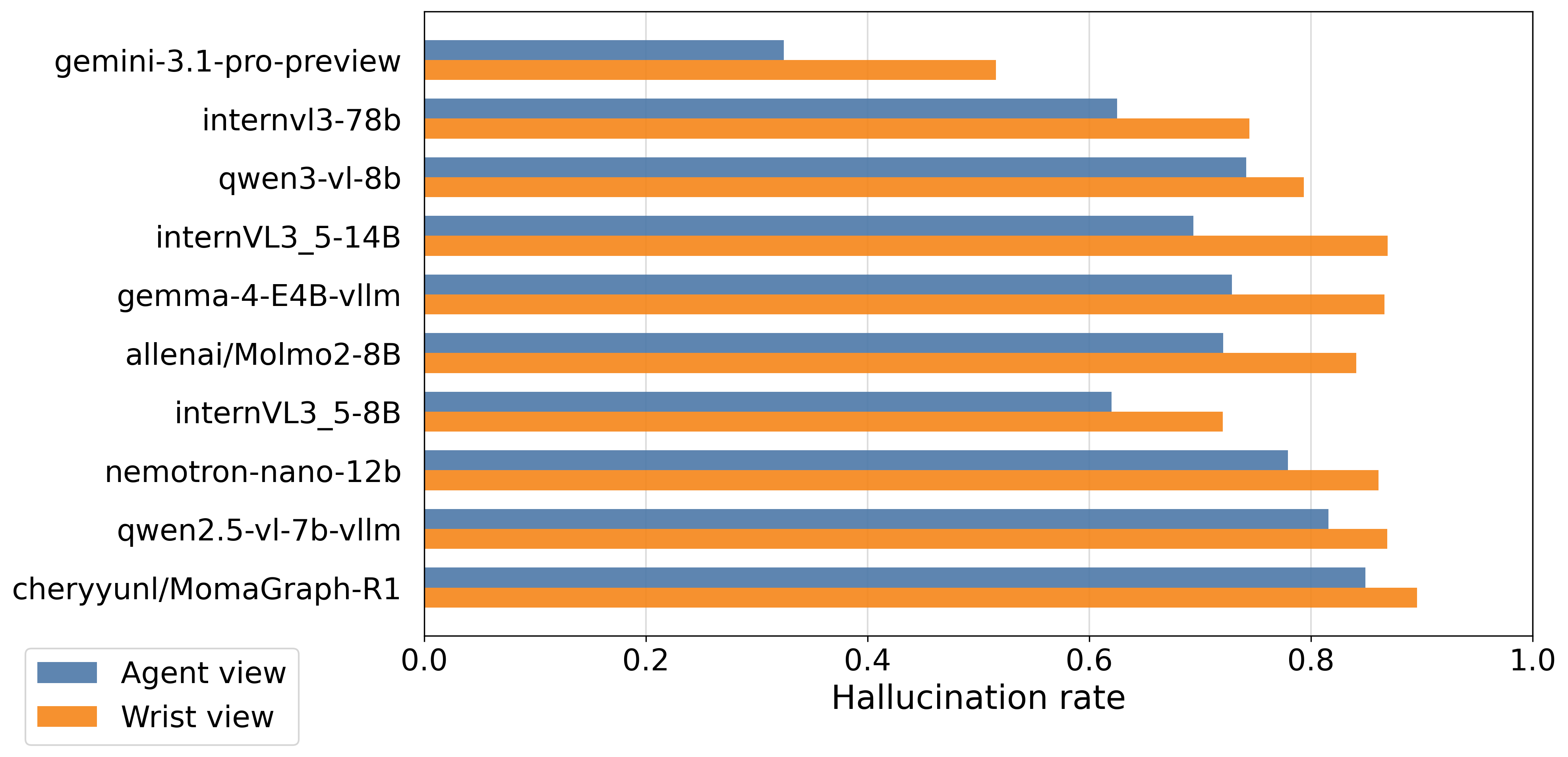}
    \includegraphics[width=0.48\linewidth]{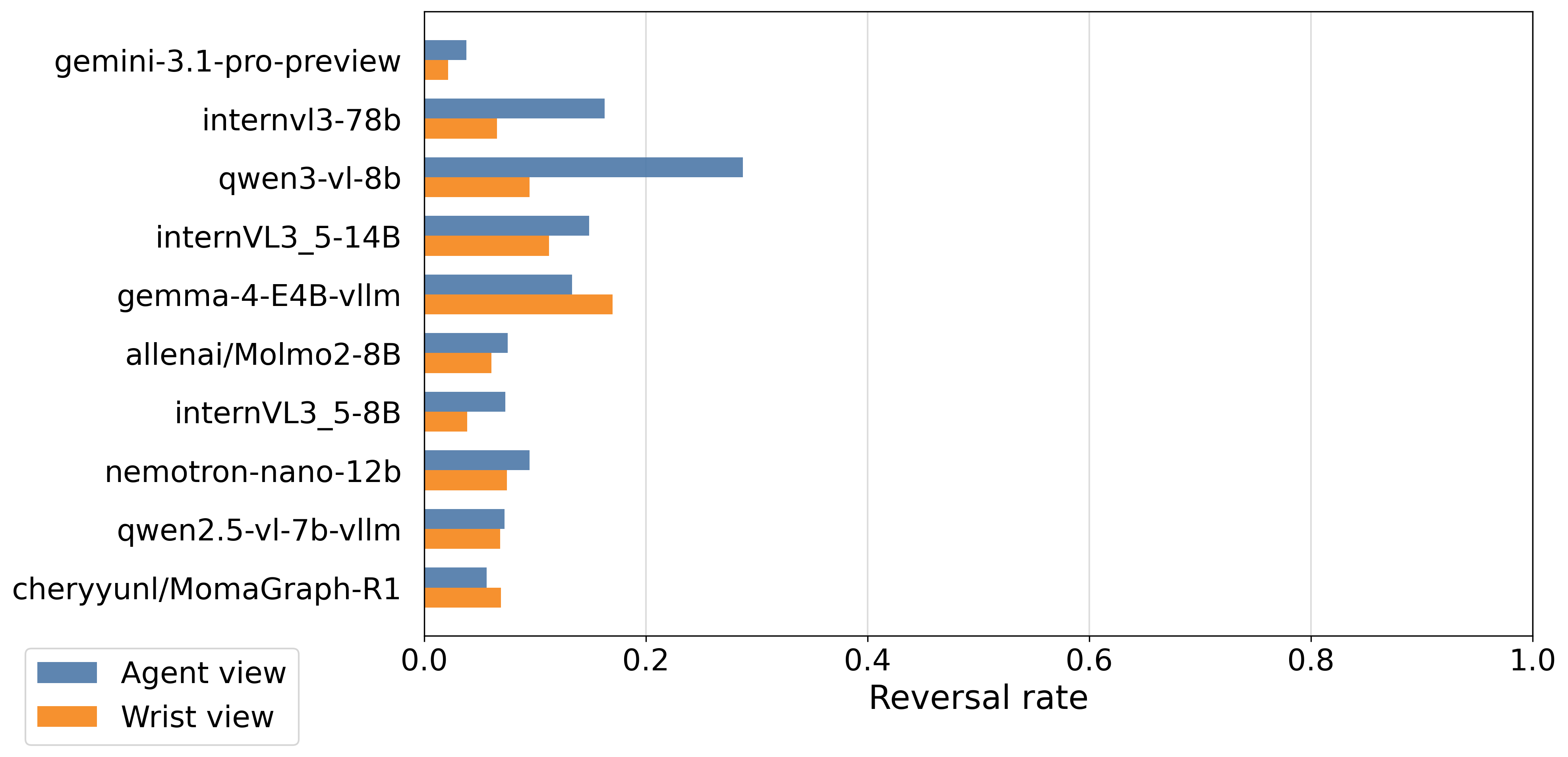}
    \includegraphics[width=0.48\linewidth]{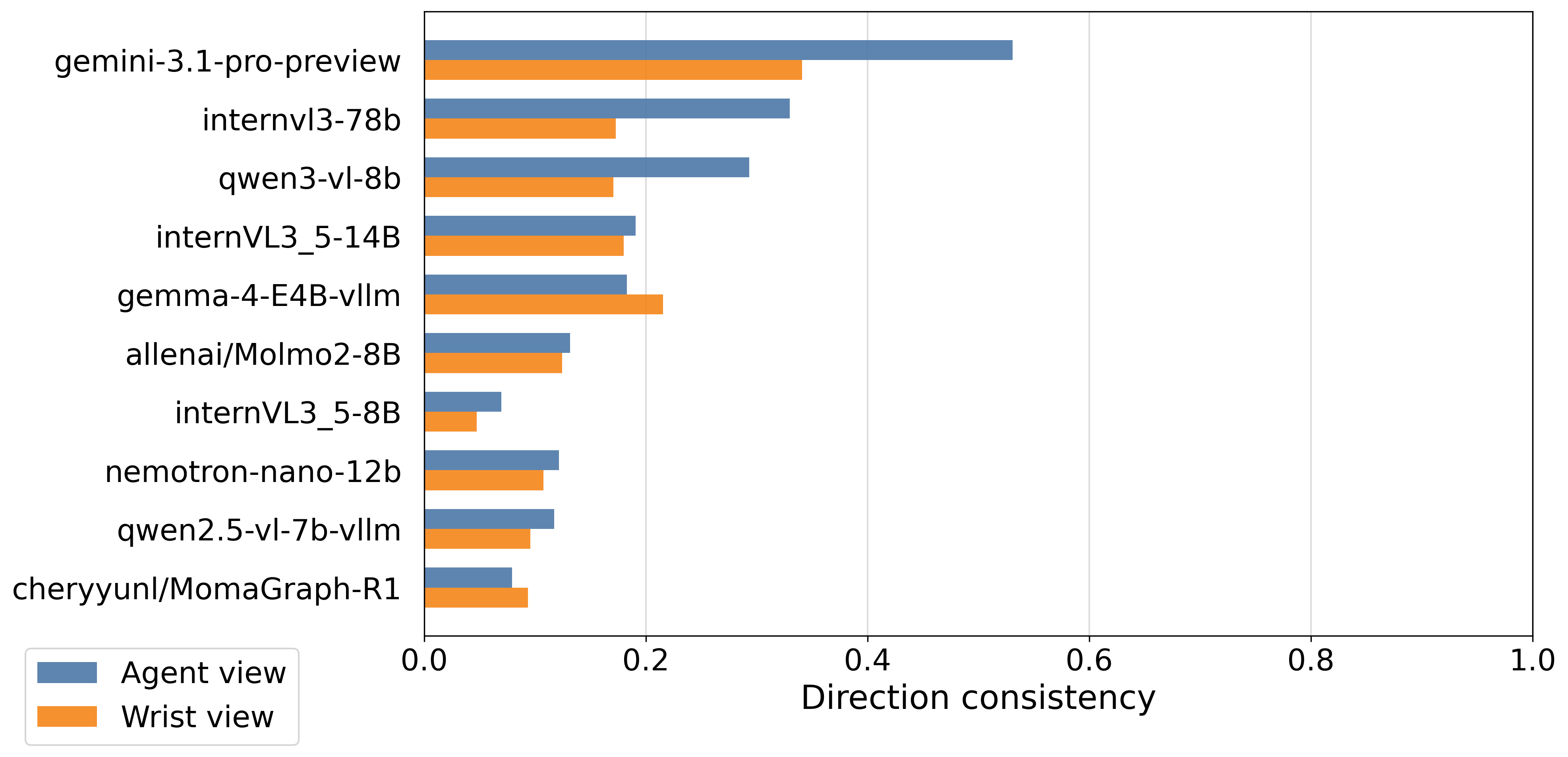}
    \caption{Diagnostic metrics for VLM scene-graph prediction.}
    \label{fig:app-vlm-diagnostics}
\end{figure}

\begin{figure}[p]
    \centering
    \includegraphics[width=0.48\linewidth]{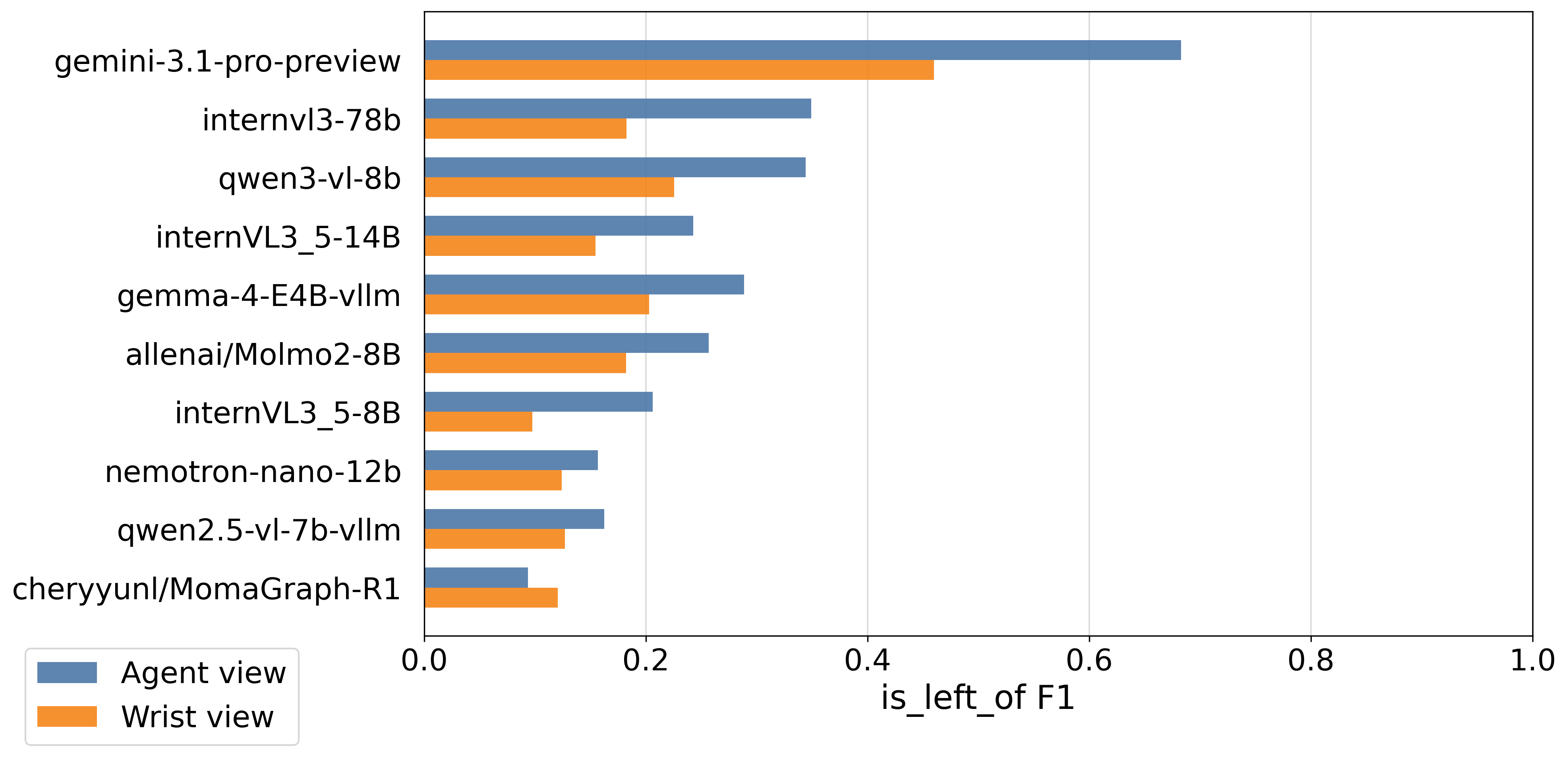}
    \includegraphics[width=0.48\linewidth]{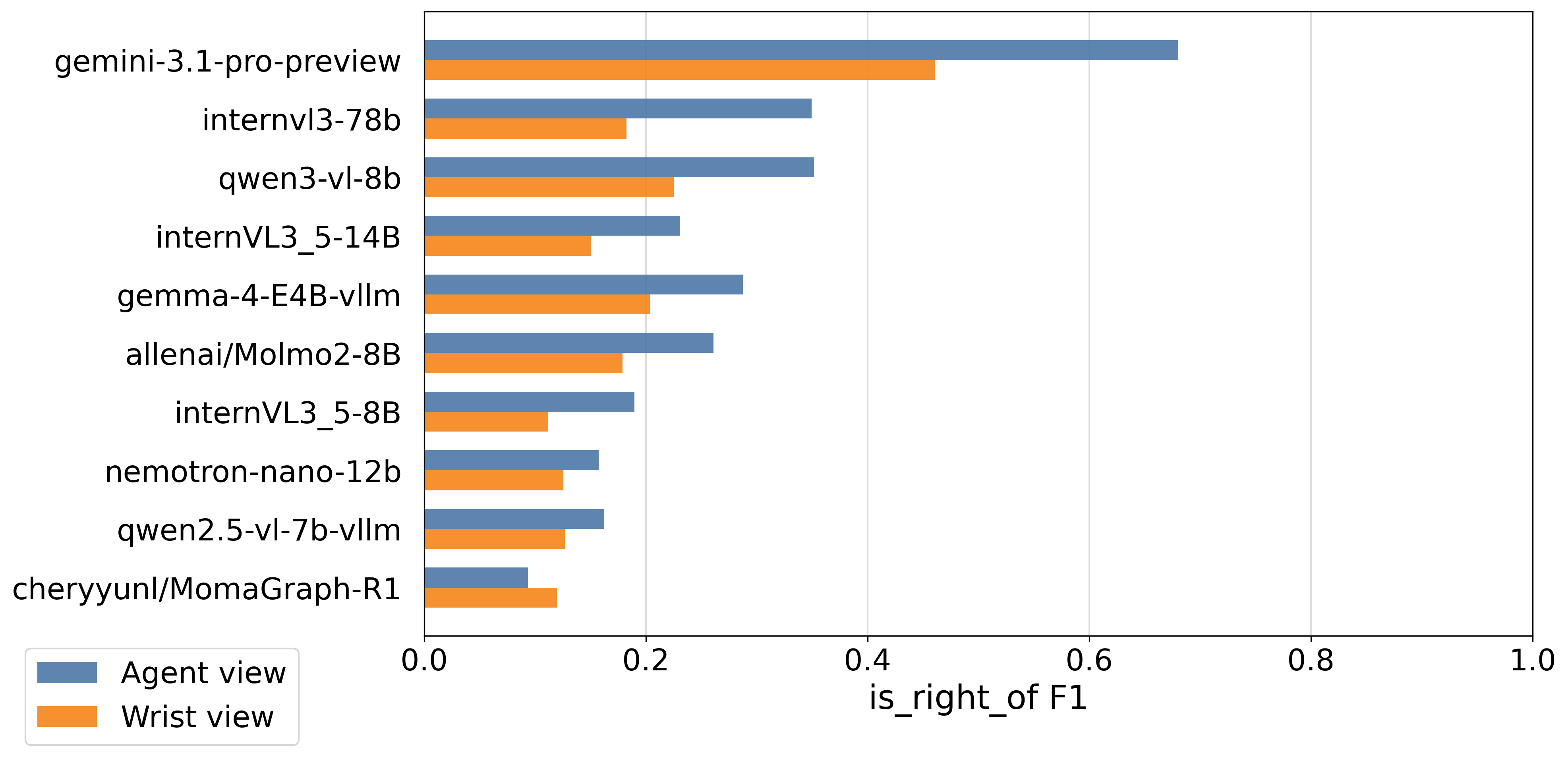}
    \includegraphics[width=0.48\linewidth]{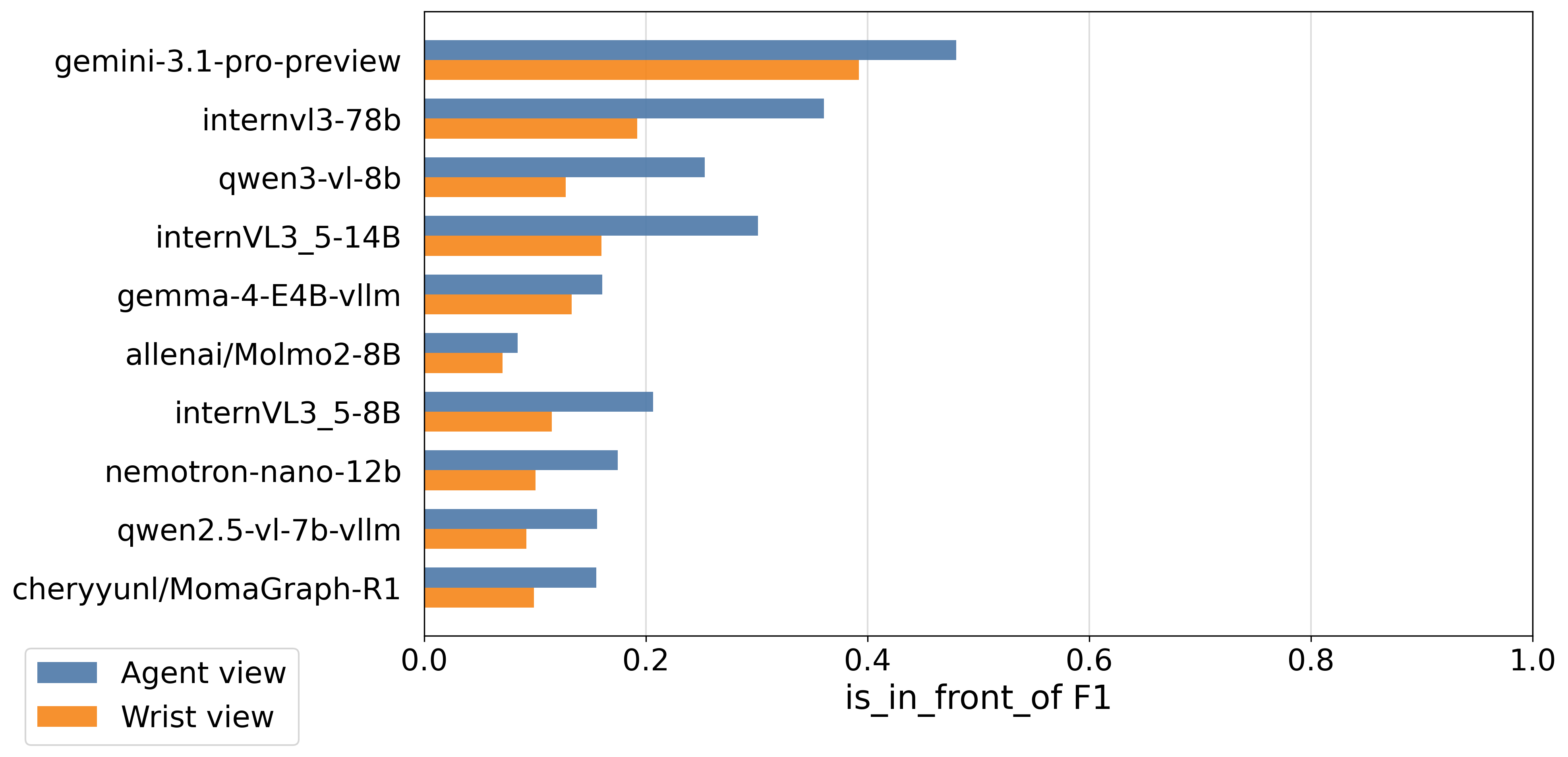}
    \includegraphics[width=0.48\linewidth]{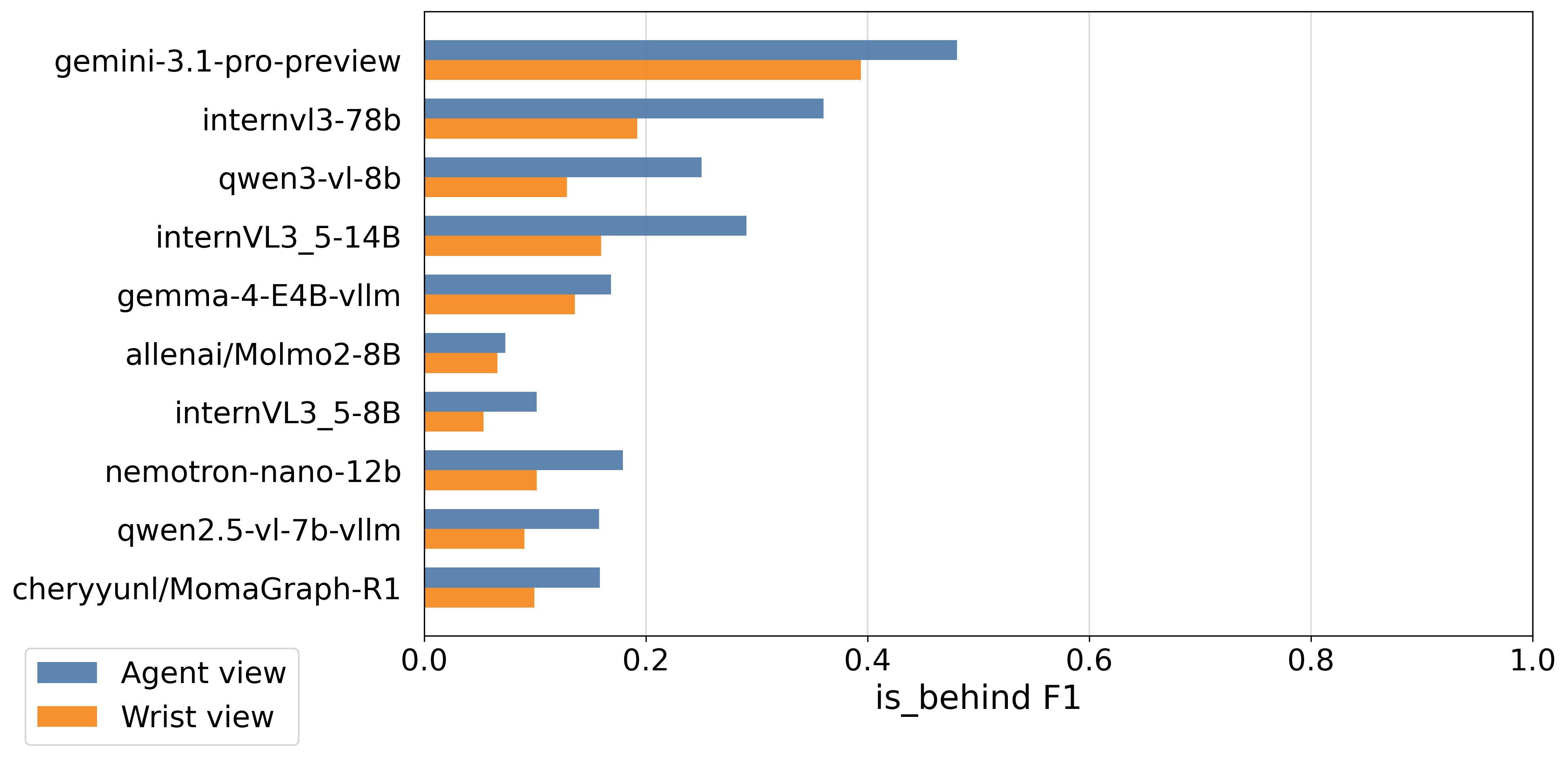}
    \caption{Per-relation F1 for lateral and depth relations.}
    \label{fig:app-vlm-horizontal-depth-relations}
\end{figure}

\begin{figure}[p]
    \centering
    \includegraphics[width=0.48\linewidth]{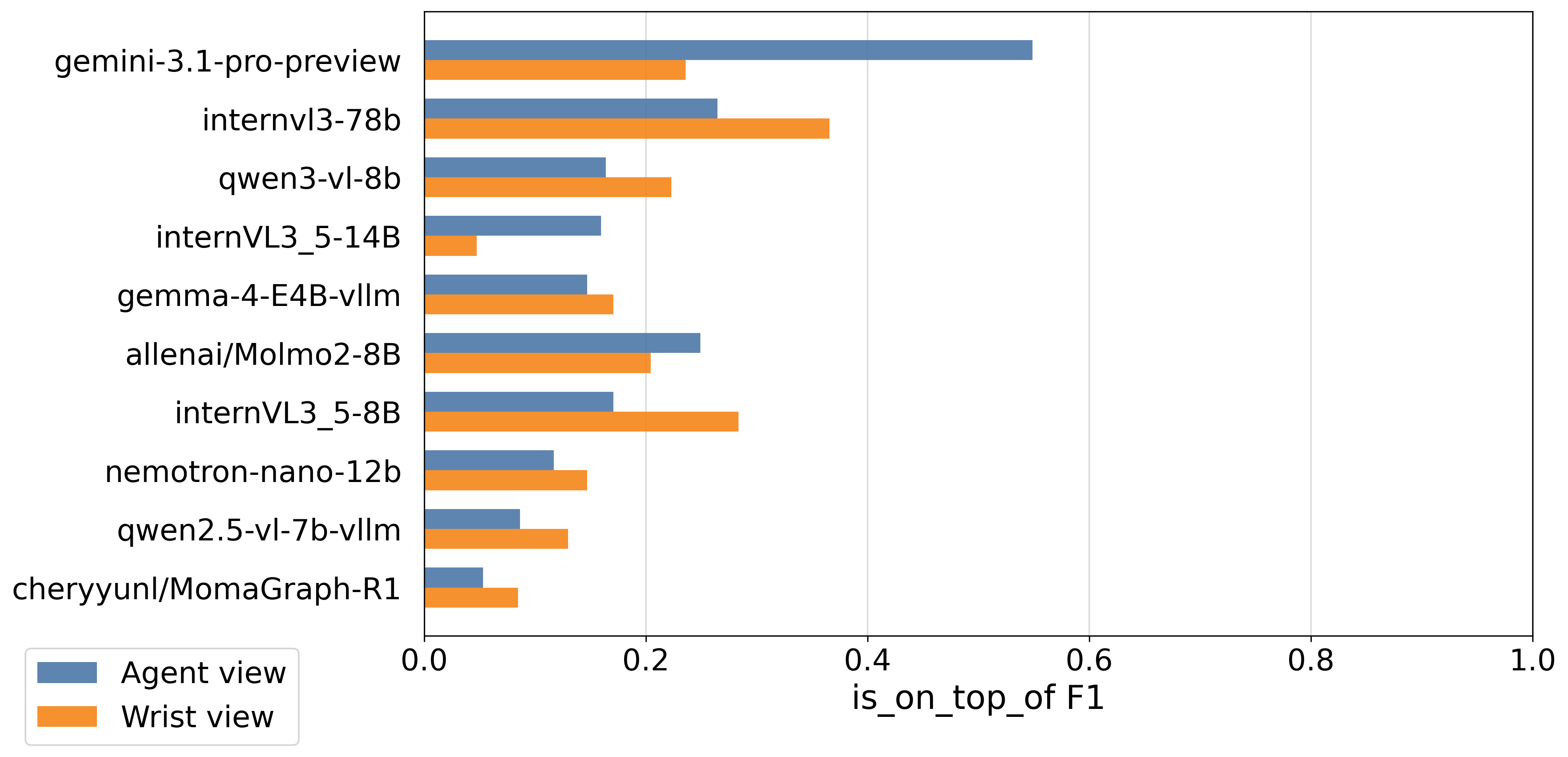}
    \includegraphics[width=0.48\linewidth]{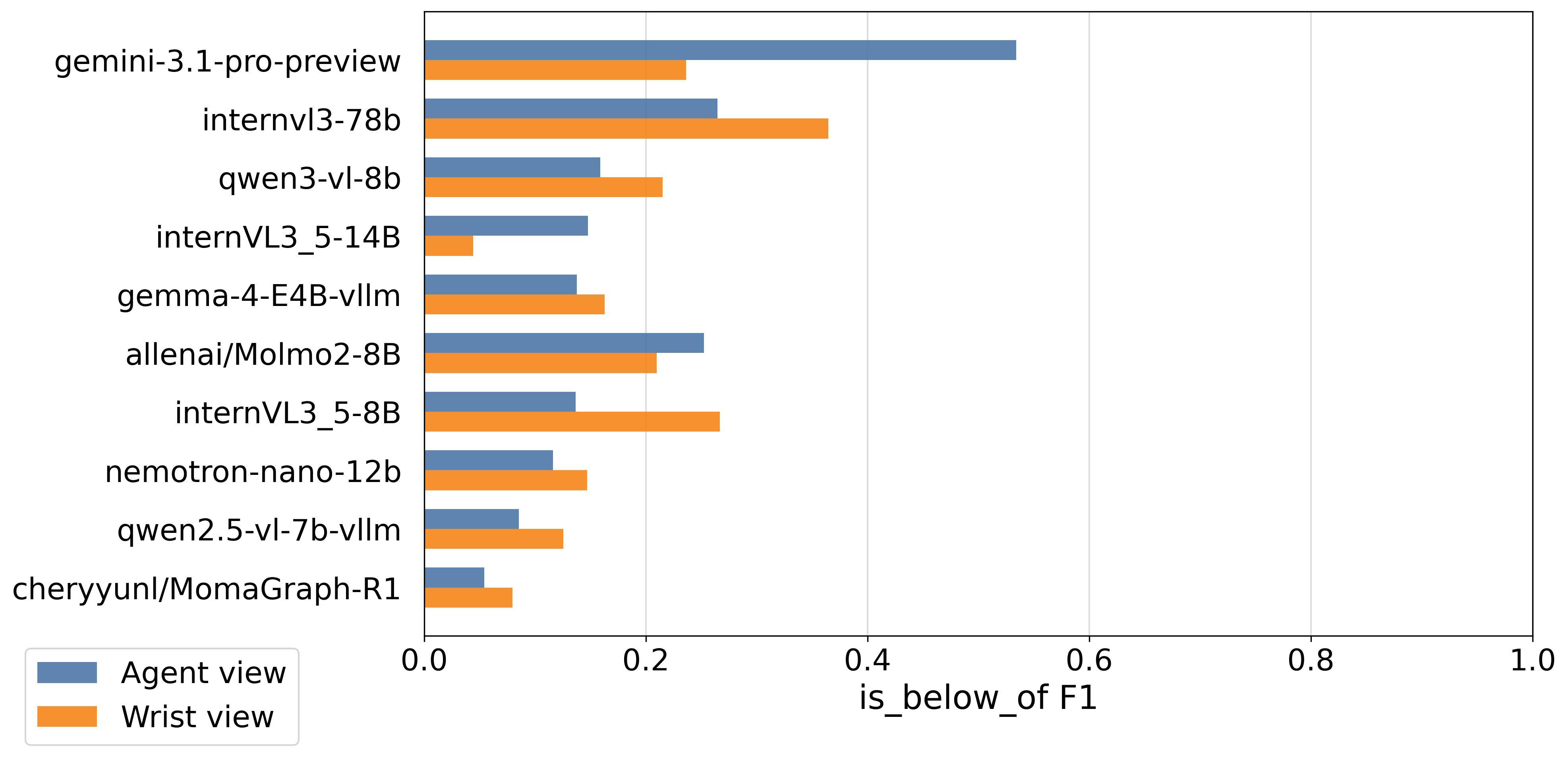}
    \includegraphics[width=0.48\linewidth]{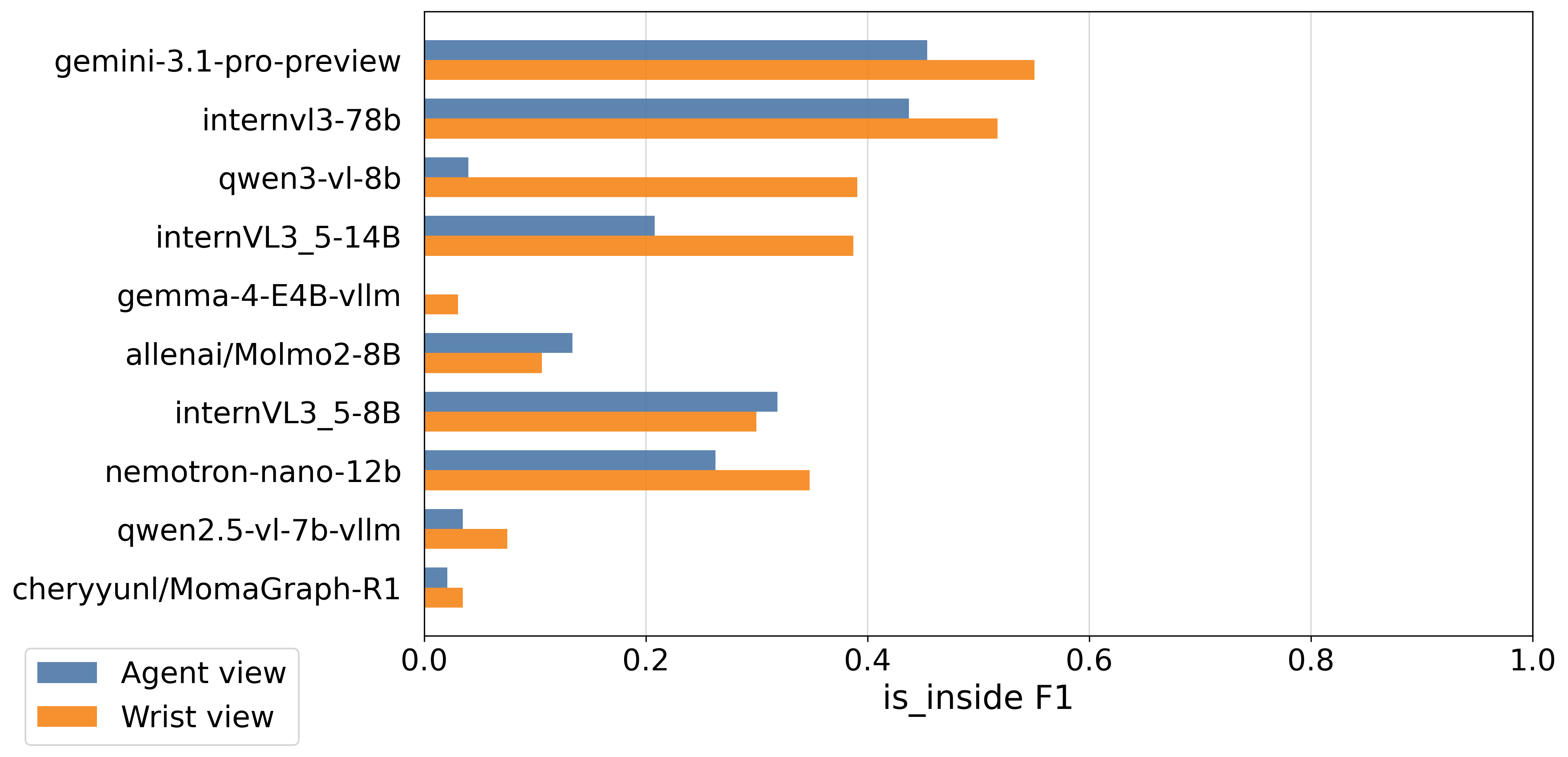}
    \includegraphics[width=0.48\linewidth]{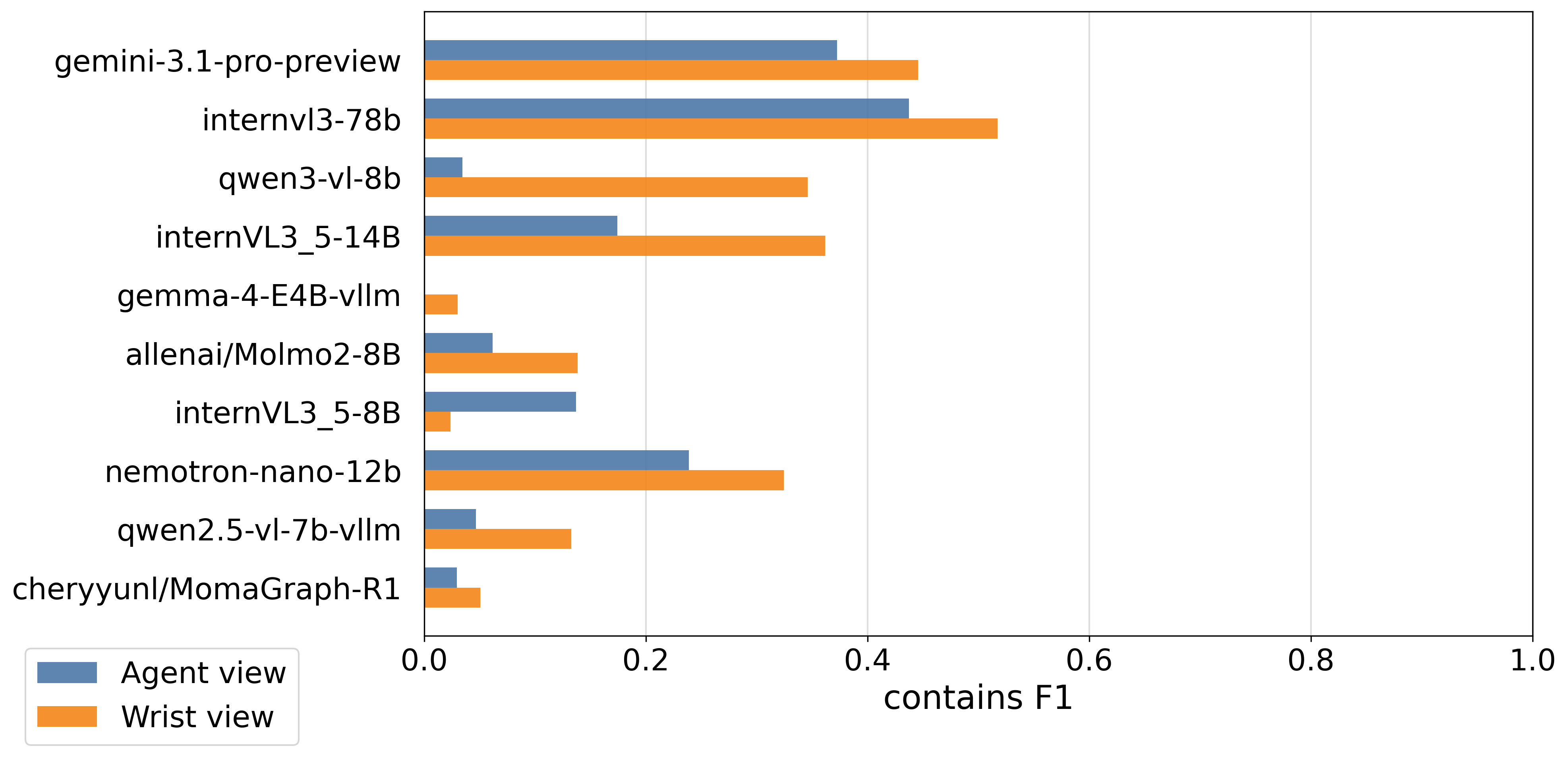}
    \caption{Per-relation F1 for support and containment relations.}
    \label{fig:app-vlm-support-containment-relations}
\end{figure}

\end{document}